\documentclass[11pt,a4paper]{article}
\usepackage{times,latexsym}
\usepackage{url}
\usepackage[T1]{fontenc}

\usepackage[acceptedWithA]{tacl}
\usepackage{times}
\usepackage{relsize}
\usepackage{latexsym}
\usepackage{xspace}
\usepackage{threeparttable}
\usepackage{booktabs}
\usepackage{graphicx}
\usepackage{framed}
\usepackage{caption}
\usepackage{adjustbox}
\usepackage{lipsum}
\usepackage{enumitem}
\usepackage{ragged2e,array,booktabs}
\usepackage{pbox}
\usepackage{dsfont}
\usepackage{graphicx}
\usepackage{textcomp}
\usepackage{multirow}
\usepackage{pgfplots}
\usepackage{paralist}
\usepackage{pbox}
\usepackage{diagbox}
\usepackage{mathtools}
\usepackage{amsmath}
\usepackage{amssymb}
\usepackage{pbox}
\usepackage{etoolbox}

\usepackage{ascii}
\usepackage{newunicodechar}
\usepackage{fdsymbol}

\usetikzlibrary{positioning}
\usetikzlibrary{calc}
\usetikzlibrary{fit}
\usetikzlibrary{decorations.text}

\usepackage{xspace,mfirstuc,tabulary}

\newcommand{\ex}[1]{\textit{#1}}

\newif\iftaclinstructions
\taclinstructionsfalse 
\iftaclinstructions
\renewcommand{\confidential}{}
\renewcommand{\anonsubtext}{(No author info supplied here, for consistency with
TACL-submission anonymization requirements)}
\newcommand{\instr}
\fi

\iftaclpubformat 

\else

\fi

\newcommand{\dataset}[1]{#1\xspace}
\newcommand{\wikipedia}{\dataset{Wikipedia}}
\newcommand{\wiki}{\dataset{Wiki}}
\newcommand{\cnn}{\dataset{CNN}}
\newcommand{\To}{\ensuremath{\rightarrow}}
\newcommand{\injected}{\dataset{INSteD}}

\newcommand{\z}{\phantom{0}}
\newcommand{\method}[2][]{#2#1\xspace}
\newcommand{\bow}{\method{BoW}}

\newcommand{\bilstm}{\method{Bi-LSTM}}
\newcommand{\infersent}{\method{InferSent}}
\newcommand{\skipthought}{\method{Skip-Thought}}
\newcommand{\bert}[1][]{\method[#1]{BERT}}
\newcommand{\bertbase}[1][]{\method[#1]{BERT-Base}}
\newcommand{\bertlarge}[1][]{\method[#1]{BERT-Large}}
\newcommand{\xlnet}[1][]{\method[#1]{XLNet}}
\newcommand{\xlnetbase}[1][]{\method[#1]{XLNet-Base}}
\newcommand{\xlnetlarge}[1][]{\method[#1]{XLNet-Large}}
\newcommand{\roberta}[1][]{\method[#1]{RoBERTa}}
\newcommand{\robertabase}[1][]{\method[#1]{RoBERTa-Base}}
\newcommand{\robertalarge}[1][]{\method[#1]{RoBERTa-Large}}
\newcommand{\albert}[1][]{\method[#1]{ALBERT}}
\newcommand{\albertlarge}[1][]{\method[#1]{ALBERT-Large}}
\newcommand{\albertxxlarge}[1][]{\method[#1]{ALBERT-xxLarge}}
\newcommand{\majority}[1][]{\method[#1]{Majority-class}}

\newcommand{\albertxxlargefreeze}[1][]{\method[#1]{ALBERT-xxLarge-freeze}}

\newcommand{\CNNtrained}{$_{\text{CNN}}$}

\newcommand{\gender}{\emph{Gender}\xspace}
\newcommand{\animacy}{\emph{Animacy}}
\newcommand{\demonstrative}{\emph{Demonstrative}\xspace}
\newcommand{\conjunction}{\emph{Conjunction}\xspace}
\newcommand{\past}{\emph{Past to Future}\xspace}
\newcommand{\negation}{\emph{Negation}\xspace}
\newcommand{\numbermanipulation}{\emph{Number}\xspace}
\newcolumntype{C}[1]{>{\centering\let\newline\\\arraybackslash\hspace{0pt}}m{#1}}
\newcolumntype{R}[1]{>{\raggedleft\let\newline\\\arraybackslash\hspace{0pt}}m{#1}}

\newcommand{\Acc}{Acc\xspace}
\newcommand{\Fscore}{F$_1$\xspace}

\newcommand{\deltapos}{\ensuremath{+}\xspace}
\newcommand{\deltaneg}{\ensuremath{-}\xspace}

\newcommand{\example}[1]{\textit{#1}}

\newcommand{\figref}[2][]{Figure#1~\ref{#2}\xspace}
\newcommand{\tabref}[2][]{Table#1~\ref{#2}\xspace}
\newcommand{\secref}[1]{Section~\ref{#1}\xspace}

\newcommand{\eg}{e.g.\xspace}
\newcommand{\ie}{i.e.\xspace}

\newcommand{\B}[1]{\textbf{#1}}

\newcolumntype{P}[1]{>{\centering\arraybackslash}p{#1}}


\title{Evaluating Document Coherence Modelling}

\author{
 Aili Shen$^\clubsuit$, Meladel Mistica$^\clubsuit$, Bahar Salehi$^\clubsuit$,  \\ \textbf{Hang Li$^\diamondsuit$, Timothy Baldwin$^\clubsuit$, Jianzhong Qi$^\clubsuit$} \\
 $^{\clubsuit}$ The University of Melbourne \\
 $^{\diamondsuit}$ AI Lab at ByteDance \\
  {\sf \{aili.shen, misticam, tbaldwin, jianzhong.qi\}@unimelb.edu.au } \\
  {\sf baharsalehi@gmail.com, lihang.lh@bytedance.com}
}

\date{}

\begin{document}
\maketitle
\begin{abstract}

While pretrained language models (``LMs'') have driven impressive gains over morpho-syntactic and semantic tasks, their ability to model discourse and pragmatic phenomena is less clear. As a step towards a better understanding of their discourse modelling capabilities, we propose a sentence intrusion detection task. We examine the performance of a broad range of pretrained LMs on this detection task for English. Lacking a dataset for the task, we introduce \injected, a novel \textbf{\underline{in}}truder \textbf{\underline{s}}en\textbf{\underline{te}}nce \textbf{\underline{d}}etection dataset, containing 170,000+ documents constructed from English \wikipedia and \cnn news articles. Our experiments show that pretrained LMs perform impressively in in-domain evaluation, but experience a substantial drop in the cross-domain setting, indicating limited generalisation capacity. Further results over a novel linguistic probe dataset show that there is substantial room for improvement, especially in the cross-domain setting.
\end{abstract}

\section{Introduction}
\label{introduction}

Rhetorical relations refer to the transition of one sentence to the next in a span of text~\cite{Mann:88,Asher:03}. They are important as a discourse device that contributes to the overall coherence, understanding, and flow of the text. These relations span a tremendous breadth of types, including \emph{contrast}, \emph{elaboration}, \emph{narration}, and \emph{justification}. These connections allow us to communicate cooperatively in understanding one another~\cite{Grice:75,Wilson:04}. 
The ability to understand such coherence (and conversely detect incoherence) is potentially beneficial for downstream tasks, such as 
storytelling \cite{Fan:19,Hu:20b}, recipe generation \cite{Chandu:19}, document-level text generation \cite{Park:15,Holtzman:18}, 
and essay scoring \cite{Tay:18,Li:18}.

However, there is little work on document coherence understanding, especially 
examining the capacity of pretrained LMs to model the coherence of longer documents. To
address this gap, we examine the capacity of pretrained language
models to capture document coherence, focused around two
research questions: 
(1) do models truly capture the intrinsic properties of document
coherence? and (2) what types of document incoherence can/can't these
models detect?

We propose the sentence intrusion detection task: (1) to determine whether a document contains an intruder sentence (coarse-grained level); and (2) to identify the span of any intruder sentence (fine-grained level). We restrict the scope of the intruder text to a single sentence, noting that in practice, the incoherent text could span multiple sentences, or alternatively be sub-sentential.

Existing datasets in document coherence measurement
\cite{Chen:19,Clercq:14,Lai:18,Mim:19,Pitler:08,Nguyen:17} are
unsuitable for our task: they are either prohibitively small, or do not specify the span of incoherent text. 
For example, in the dataset of \citet{Lai:18}, each document is assigned
a coherence score, but the span of incoherent text is not
specified. There is thus a need for a large-scale dataset which includes
annotation of the position of intruder text. Identifying the span of incoherent text can benefit tasks where explainability and immediate feedback are important, such as essay scoring \cite{Tay:18,Li:18}. 

In this work, we introduce a dataset consisting of English documents
from two domains: \wikipedia articles (106K) and CNN news articles
(72K). This dataset fills a gap in research pertaining to document
coherence: our dataset is large in scale, includes
both coherent and incoherent documents, and has mark-up of the position
of any intruder sentence. \figref{injected_document_example} is an
example document with an intruder sentence. Here, the highlighted
sentence reads as though it should be an elaboration of the previous
sentence, but clearly exhibits an abrupt change of
topic and the pronoun \ex{it} cannot be readily resolved.

\begin{figure}[!t]
  \small
	\noindent\begin{framed}
		(1) Mark Ferguson (born 21 May 1990) is an Irish handballer, currently playing in Dublin, Ireland.
		
		(2) \textbf{It is a twelve time Asian Champion, the tournament has been won by any other nation only twice.}
		
		(3) Previously playing for his university team ITB, in the 2013/14 League his club Lughnasa HC came 3rd ...
		
		(4) Mark has been involved in the National Team from 2011 and played in Ireland's first ever European qualifiers ...
		
		(5) The following year Ireland took part in the 2016 Men's European Championship Qualification ...
	\end{framed}
	\caption{An excerpt of an incoherent document, with the ``intruder'' sentence indicated in \textbf{bold}. }
	\label{injected_document_example}
\end{figure}

This paper makes the following contributions: (1) we propose the
sentence intrusion detection task, and examine how pretrained LMs
perform over the task and hence at document coherence understanding; (2)
we construct a large-scale dataset from two domains --- \wikipedia and
CNN news articles --- that consists of coherent and incoherent
documents, and is accompanied with the positions of intruder sentences,
to evaluate in both in-domain and cross-domain settings; (3) we examine
the behaviour of models and humans, to better understand the ability of models to model the intrinsic properties of document coherence; and (4) we
further hand-craft adversarial test instances across a variety of
linguistic phenomena to better understand the types of incoherence that
a given model can detect.

\section{Related Work}
\label{related_work}

We first review tasks relevant to our proposed task, then describe existing datasets used in coherence measurement, and finally discuss work on dataset artefacts and linguistic probes.

\subsection{Document Coherence Measurement}

Coherence measurement has been studied across various tasks, such as the {document discrimination task} \cite{Barzilay:05,Elsner:07, Barzilay:08,Elsner:11,Li:17,Putra:17}, {sentence insertion} \cite{Elsner:11,Putra:17,Xu:19}, {paragraph reconstruction} \cite{Lapata:03,Elsner:07,Li:17,Xu:19,Prabhumoye:20}, {summary coherence rating} \cite{Barzilay:05,Pitler:10,Guinaudeau:13,Nguyen:17}, {readability assessment} \cite{Guinaudeau:13,Mesgar:16,Mesgar:18}, and {essay scoring} \cite{Mesgar:18,Somasundaran:14,Tay:18}. These tasks differ from our task of intruder sentence detection as follows. First, the document discrimination task assigns coherence scores to a document and its sentence-permuted versions, where the original document is considered to be well-written and coherent and permuted versions incoherent. Incoherence is introduced by shuffling sentences, while our intruder sentences are selected from a second document, and there is only ever a single intruder sentence per document. Second, sentence insertion aims to find the correct position to insert a removed sentence back into a document. Paragraph reconstruction aims to recover the original sentence order of a shuffled paragraph given its first sentence. These two tasks do not consider sentences from outside of the document of interest. Third, the aforementioned three tasks are artificial, and have very limited utility in terms of real-world tasks, while our task can provide direct benefit in applications such as essay scoring, in identifying incoherent (intruder) sentences as a means of providing user feedback and explainability of essay scores. Lastly, in summary coherence rating, readability assessment, and essay scoring, coherence is just one dimension of the overall document quality measurement.

Various methods have been proposed to capture local and global coherence, while our work aims to examine the performance of existing pretrained LMs in document coherence understanding. To assess local coherence, traditional studies have used entity matrices, \eg to represent entity transitions across sentences \citep{Barzilay:05,Barzilay:08}. 
\citet{Guinaudeau:13} and \citet{Mesgar:16} use a graph to model entity transition sequences. Sentences in a document are represented by nodes in the graph, and two nodes are connected if they share the same or similar entities. Neural models have also been proposed~\cite{Ji:17,Li:17,Li:18,Mesgar:18,Mim:19,Nguyen:17}. For example, 
\citet{Tay:18} capture local coherence by computing the similarity of the output of two LSTMs \citep{Hochreiter:97}, which they concatenate with essay representations to score essays. \citet{Li:18} use multi-headed self-attention to capture long distance relationships between words, which are passed to an LSTM layer to estimate essay coherence scores. \citet{Xu:19} use the average of local coherence scores between consecutive pairs of sentences as the document coherence score.

Another relevant task is disfluency detection in spontaneous speech transcription \citep{Johnson:04,Lou:18}. This task detects the {reparandum} and {repair} in spontaneous speech transcriptions to make the text fluent by replacing the {reparandum} with the {repair}. Also relevant is language identification in code-switched text \cite{Adouane:18a,Adouane:18b,Mave:18,Yirmibesoglu:18}, where disfluency is defined at the language level (for a monolingual speaker, \eg). \citet{Lau:15} and \citet{Warstadt:19} predict sentence-level acceptability (how natural a sentence is). However, none of tasks are designed to measure document coherence, although sentence-level phenomena can certainly impact on document coherence.

\subsection{Document Coherence Datasets}

There exist a number of datasets targeted at discourse understanding. For example, \citet{Alikhani:19} construct a multi-modal dataset for understanding discourse relations between text and imagery, such as elaboration and exemplification. In contrast, we focus on discourse relations in a document at the inter-sentential level. The {Penn Discourse Treebank} \cite{Miltsakaki:04, Prasad:08} is a corpus of coherent documents with annotations of discourse connectives and their arguments, noting that inter-sentential discourse relations are not always lexically marked \cite{Webber:09}. 

The most relevant work to ours is the discourse coherence dataset of \citet{Chen:19}, which was proposed to evaluate the capabilities of pretrained LMs in capturing discourse context. This dataset contains documents (18K \wikipedia articles and 10K documents from the Ubuntu IRC channel) with fixed sentence length, and labels documents only in terms of whether they are incoherent, without considering the position of the incoherent sentence. In contrast, our dataset: (1) provides more fine-grained information (\ie the sentence position); (2) is larger in scale (over 170K documents); (3) contains documents of varying length; (4) incorporates adversarial filtering to reduce dataset artefacts (see \secref{dataset_construction}); and (5) is accompanied with human annotation over the \wikipedia subset, allowing us to understand behaviour patterns of machines and humans. 

\subsection{Dataset Artefacts}
\label{sec:artefacts}

Also relevant to this research is work on removing artefacts in datasets \cite{Zellers:19,McCoy:19,Zellers:18}. For example, based on analysis of the {SWAG} dataset \cite{Zellers:18}, \citet{Zellers:19} find artefacts such as stylistic biases, which correlate with the document labelling and mean that naive models are able to achieve abnormally high results. Similarly, \citet{McCoy:19} examine artefacts in an NLI dataset, and find that naive heuristics which are not directly related to the task can perform remarkably well. We incorporate the findings of such work in the construction of our dataset.

\subsection{Linguistic Probes}
\label{sec:linguistic_probe}

Adversarial training has been used to craft adversarial examples to obtain more robust models, either by manipulating model parameters (white-box attacks) or minimally editing text at the character/word/phrase level (black-box attacks). For example, \citet{Papernot:18} provide a reference library of adversarial example construction techniques and adversarial training methods. 

As we aim to understand the linguistic properties that each model has captured, we focus on black-box attacks \cite{Sato:18,Cheng:18,Liang:18,Yang:18b,Samanta:17}. For example, \citet{Samanta:17} construct adversarial examples for sentiment classification and gender detection by deleting, replacing, or inserting words in the text. For a comprehensive review of such studies, see \citet{Belinkov:19}.

There is also a rich literature on exploring what kinds of linguistic phenomena a model has learned \cite{Hu:20,Hewitt:19a,Hewitt:19b,Chen:19,McCoy:19,Baroni:18,Gulordava:18,Peters:18,Tang:18,Blevins:18,Wilcox:18,Clark:18,Tran:18,Belinkov:17}. The basic idea is to use learned representations to predict linguistic properties of interest. Example linguistic properties are subject--verb agreement or syntactic structure, while representations can be word or sentence embeddings. For example, \citet{Marvin:18} construct minimal sentence pairs, consisting of a grammatical and ungrammtical sentence, to explore the capacity of LMs in capturing phenomena such as subject--verb agreement, reflexive anaphora, and negative polarity items. 
In our work, we hand-construct intruder sentences which result in incoherent documents, based on a broad range of linguistic phenomena.

\section{Dataset Construction}
\label{dataset_construction}

\subsection{Dataset Desiderata}
\label{desiderata}

To construct a large-scale, low-noise dataset that truly tests the ability of systems to detect intruder sentences, we posit five desiderata:

\begin{compactenum}
\item \textbf{Multiple sources:} The dataset should not be too homogeneous in terms of genre or domain, and should ideally test the ability of models to generalise across domain.
\item \textbf{Defences against hacking:} Human annotators and machines should not be able to hack the task and reverse-engineer the labels by sourcing the original documents.

\item \textbf{Free of artefacts:} The dataset should be free of artefacts, that allow naive heuristics to perform well.

\item \textbf{Topic consistency:} The intruder sentence, which is used to replace a sentence from a coherent document to obtain an incoherent document, should be relevant to the topic of the document, to focus the task on coherence and not simple topic detection.

\item \textbf{KB-free:} Our goal is \emph{NOT} to construct a fact-checking dataset; the intruder sentence should be determinable based on the content of the document, without reliance on external knowledge bases or fact-checking. 

\end{compactenum}

\subsection{Data Sources}
\label{data_sources}

We construct a dataset from two sources --- \wikipedia and \cnn~--- which differ in style and genre, satisfying the first desideratum. Similar to {WikiQA} \cite{Yang:15} and {HotpotQA} \cite{Yang:18}, we represent a \wikipedia document by its summary section (\ie, the opening paragraph), constraining the length to be between 3 and 8 sentences. For \cnn, we adopt the dataset of \citet{Hermann:15} and \citet{Nallapati:16}, which consists of over 100,000 news articles. To obtain documents with sentence length similar to those from \wikipedia, we randomly select the first 3--8 sentences from each article.

To defend against dataset hacks\footnote{Deliberate or otherwise, \eg via pre-training on the same version of Wikipedia our dataset was constructed over.} which could expose the labels of the test data (desideratum 2), the \wikipedia test set is randomly sampled from 37 historical dumps of \wikipedia, where the selected article has a cosine similarity less than the historical average of 0.72 with its online version.\footnote{This threshold was determined by calculating the average {TF-IDF}-weighted similarity of the summary section for documents in all 37 dumps with their current online versions.} For the training set, we remove this requirement and randomly select articles from different \wikipedia dumps, \ie, the articles in the training set might be the same as their current online version. For \cnn, we impose no such limitations.

\subsection{Generating Candidate Positive Samples}
\label{positive_sample_generation}

We consider the original documents to be coherent. We construct incoherent documents from half of our sampled documents as follows (satisfying desiderata 3--5):

\begin{compactenum}
	\item Given a document $D$, use bigram hashing and TF-IDF matching \cite{Chen:17} to retrieve the top-$10$ most similar documents from a collection of documents from the same domain, where $D$ is the query text. Let the set of retrieved documents be $\mathcal{R}_D$.
	\item Randomly choose a non-opening sentence $S$ from document $D$, to be replaced by a sentence candidate generated later. We do not replace the opening sentence as it is needed to establish document context.
	\item For each document $D' \in \mathcal{R}_D$, randomly select one non-opening sentence $S' \in D'$ as an intruder sentence candidate. 
	\item Calculate the TF-IDF-weighted cosine similarity between sentence $S$ and each candidate 
	$S'$. Remove any candidates with similarity scores $\ge 0.6$, to attempt to generate a KB-free incoherence.
    \item Replace sentence $S$ with each low-similarity candidate $S'$, and use a fine-tuned \xlnetlarge model \citep{Yang:19} to check whether it is easy for \xlnetlarge to detect (see \secref{sec:models}). For documents with both easy and difficult sentence candidates, we randomly sample from the difficult sentence candidates; otherwise, we randomly choose from all the sentence candidates.
\end{compactenum}

The decision to filter out sentence candidates with similarity $\ge 0.6$ was based on the observation that more similar sentences often led to the need for world knowledge to identify the intruder sentence (violating the fifth desideratum). 
For example, given \example{It is the \textbf{second} novel in the first of three trilogies about Bernard Samson, ...}, a candidate intruder sentence candidate with high similarity is \example{It is the \textbf{first} novel in the first of three trilogies about Bernard Samson ...}. 

We also trialled other ways of generating incoherent samples, such as using sentence $S$ from document $D$ as the query text to retrieve documents, and adopting a $2$-hop process to retrieve relevant documents. We found that these methods resulted in documents which can be identified by the pretrained models easily.

\section{Dataset Analysis}
\label{dataset_analysis}

\subsection{Statistics of the Dataset}
\begin{table}[t]
	\centering
        \small
		\begin{threeparttable}
			\begin{tabular}{lccc@{}r}  
				\toprule  
                          \multirow{2}{*}{Source}	&\multirow{2}{*}{\#docs}          &avg. & avg. \\ 
                          && \#sents & \#tokens \\\midrule
				\wikipedia 	&106,352 (46\%)        &5$\pm$1  &126$\pm$24           \\ 
				\cnn	&\hphantom{0}72,670 (49\%)   & 5$\pm$1    &134$\pm$32   \\ \bottomrule
			\end{tabular}	
	\end{threeparttable}
	\caption{Dataset statistics for \injected. Numbers in parentheses are percentages of incoherent documents.} 
	\label{statistics_of_dataset} 
\end{table}

The process described in \secref{dataset_construction} resulted in 106,352 \wikipedia documents and 72,670 \cnn documents, at an average sentence length of 5 in both cases (see \tabref{statistics_of_dataset}). The percentages of positive samples (46\% and 49\%, respectively) are slightly less than 50\% due to our data generation constraints (detailed in \secref{positive_sample_generation}), which can lead to no candidate intruder sentence $S'$ being generated for original sentence $S$. We set aside 8\% of \wikipedia (which we manually tag, as detailed in \secref{sec:verification}) and 20\% of \cnn for testing.

\begin{table*}[t!]
\small
	\centering
	\begin{tabular}{>{\raggedright}m{0.7in} >{\raggedright}m{4.9in} >{\centering\arraybackslash}m{0.08in}}
          \toprule
          Incoherence & Example & \% \\
          \midrule
          
          Information structure inconsistency & \example{He is currently the senior pastor at Sovereign Grace Church of Louisville. \textbf{The Church is led by Senior Pastor Ray Johnston, Senior Pastor Curt Harlow and Senior Pastor Andrew McCourt, and Senior Pastor Lincoln Brewster.} Under Mahaney's leadership, Sovereign Grace Church of Louisville is a member of Sovereign Grace Churches.} & 58 \\[6ex]

          Logical inconsistency & \example{Michael David, born September 22, 1954, is an American-born American painter. \textbf{From 1947--1949 he attended the Otis Art Institute, from 1947 to 1950 he also attended the Art Center College of Design in Los Angeles, and in 1950 the Chouinard Art Institute.}} &26 \\[4.5ex]

          Factual inconsistency &\example{The Newport Tower has 37 floors. \textbf{It is located on the beachfront on the east side of Collins Avenue between 68th and 69th Streets.} The building was developed by Melvin Simon \& Associates in 1990.} &35 \\ 
          \bottomrule
	\end{tabular}
	
	\caption{Types of document incoherence in \wikipedia. Text in bold indicates the intruder sentence. }
        \label{injected_type_analysis}
\end{table*}

\subsection{Types of Incoherence}

To better understand the different types of issues resulting from our automatic method, we sampled 100 (synthesised) incoherent documents from \wikipedia and manually classified the causes of incoherence according to three overlapping categories (ranked in terms of expected ease of detection): (1) information structure inconsistency (a break in information flow); (2) logical inconsistency (a logically inconsistent world state is generated, such as someone attending school before they were born); and (3) factual inconsistency (where the intruder sentence is factually incorrect). See \tabref{injected_type_analysis} for a breakdown across the categories, noting that a single document can be incoherent across multiple categories. Information structure inconsistency is the most common form of incoherence, followed by factual inconsistency. The 35\% of documents with factual inconsistencies break down into 8\% (overall) that have other types of incoherence, and 27\% that only have a factual inconsistency. This is an issue for the fifth desideratum for our dataset (see \secref{desiderata}), motivating the need for manual checking of the dataset to determine how readily the intruder sentence can be detected.\footnote{We keep these documents in the dataset, as it is beyond the scope of this work to filter these documents out.} 

\subsection{Evaluation Metrics}

We base evaluation of intruder sentence detection at both the document and sentence levels:
\begin{compactitem}
\item \textbf{document level}: \textit{Does the document contain an intruder sentence?} This is measured based on classification accuracy (\Acc), noting that the dataset is relatively balanced at the document level (see \tabref{statistics_of_dataset}). A prediction is ``correct'' if at least one sentence/none of the sentences is predicted to be an intruder.
\item \textbf{sentence level}: \textit{Is a given (non-opening) sentence an intruder sentence?} This is measured based on \Fscore, noting that most (roughly 88\%) sentences are non-intruder sentences.
\end{compactitem}

\subsection{Testing for Dataset Artefacts}
\label{artefacts}

To test for artefacts, we use \xlnetlarge \cite{Yang:19} to predict whether each non-opening sentence is an intruder sentence, \textit{in complete isolation of its containing document} (\ie, as a standalone sentence classification task). We compare the performance of \xlnetlarge with a majority-class baseline (``\majority'') which predicts all sentences to be non-intruder sentences (\ie, from the original document), where \xlnetlarge is fine-tuned over the \wikipedia/\cnn training set, and tested over the corresponding test set.

For \wikipedia, \xlnetlarge obtains an \Acc of 55.4\% (vs.\ 55.1\% for \majority) and \Fscore of 3.4\% (vs.\ 0.0\% for \majority). For \cnn, the results are 50.8\% and 1.2\%, respectively (vs.\ 51.0\% and 0.0\% resp.\ for \majority). These results suggest that the dataset does not contain obvious artefacts, at least for \xlnetlarge. We also experiment with a {TF-IDF} weighted bag-of-words logistic regression model, achieving slightly worse results than \xlnetlarge (\Acc = 55.1\%, \Fscore = 0.05\% for \wikipedia, and \Acc = 50.6\%, \Fscore = 0.3\% for \cnn).\footnote{For \robertalarge (\secref{general_performance}), there were also no obvious artefacts observed in the standalone sentence setting: \Acc = 55.7\% and \Fscore = 5.3\% over \wikipedia, and \Acc = 51.3\% and \Fscore = 4.3\% over \cnn.}

\subsection{Human Verification}
\label{sec:verification}

We performed crowdsourcing via Amazon Mechanical Turk over the \wikipedia test data to examine how humans perform over this task. Each Human Intelligence Task (HIT) contained 5 documents and was assigned to 5 workers. For each document, the task was to identify a single sentence which ``creates an incoherence or break in the content flow'', or in the case of no such sentence, ``None of the above'', indicating a coherent document. In the task instructions, workers were informed that there is at most one intruder sentence per document, and were not able to select the opening sentence. Among the 5 documents for each HIT, there was one incoherent document from the training set, which was pre-identified as being easily detectable by an author of the paper, and acts as a quality control item. 
We include documents where at least $3$ humans assign the same label as our test dataset (90.3\% of the \wikipedia test dataset), where all the results are reported over these documents, if not specified.\footnote{Different people may have different thresholds in considering a document to be incoherent, but this is beyond the scope of our work.} 
Payment was calibrated to be above Australian minimum wage.

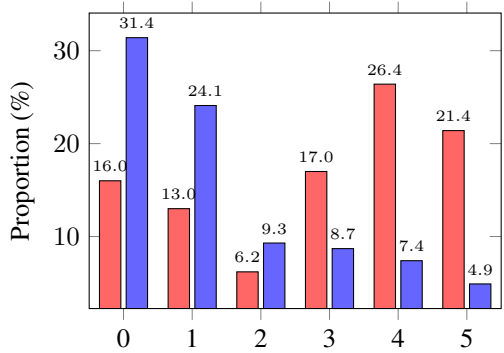
\begin{figure}[t!]
  \small
	\begin{center}
          \begin{tikzpicture}

\begin{axis}[
    ybar,bar width=8,
    width=7cm,height=5.5cm,
    enlargelimits=0.1,
    ylabel={Proportion (\%)},ylabel near ticks,
    symbolic x coords={0,1,2,3,4,5},
    xtick=data,
    nodes near coords,
    every node near coord/.append style={font=\smaller[3],
      /pgf/number format/.cd,
      fixed,
      fixed zerofill,
      precision=1,
      /tikz/.cd},
    nodes near coords align={vertical},
    ]
\addplot[fill=red!60!white] coordinates {(0,16.0) (1,13.0) (2,6.2) (3,17.0) (4,26.4) (5,21.4)};
\addplot[fill=blue!60!white] coordinates {(0,31.4) (1,24.1) (2,9.3)  (3,8.7)  (4,7.4)  (5,4.9)};
\end{axis}
\end{tikzpicture}
	\end{center}
	\caption{Distribution of instances where different numbers of humans produce correct answers. Note that the red bars indicate distributions over all documents and the blue bars indicate distributions over incoherent documents.}
	\label{human_distribution}
\end{figure}

\figref{human_distribution} shows the distribution of instances where different numbers of workers produced the correct answer (the red bar). For example, for 6.2\% of instances, 2/5 workers annotated correctly. The blue bars indicate the proportion of incoherent documents where the intruder sentence was correctly detected by the given number of annotators (\eg, for 9.3\% of incoherent documents, only 2/5 workers were able to identify the intruder sentence correctly). Humans tend to agree with each other over coherent documents, as indicated by the increasing percentages for red bars but decreasing percentages for blue bars across the $x$-axis. Intruder sentences in incoherent documents, however, are harder to detect. One possible explanation is that the identification of intruder sentences requires fact-checking, which workers were instructed not to do (and base their judgement only on the information in the provided document); another reason is that intruder sentences disrupt local incoherence with neighbouring sentences, creating confusion as to which is the intruder sentence (with many of the sentence-level mis-annotations being off-by-one errors).

\section{Models}
\label{sec:models}

We model intruder sentence detection as a binary classification task: each non-opening sentence in a document is concatenated with the document, and a model is asked to predict whether the sentence is an intruder sentence to the document. 

\begin{table*}[t]
  \small
	\centering
		\begin{threeparttable}
			\begin{tabular}{lc@{\,\,}cc@{\,\,}ccc@{\,\,}cc@{\,\,}c}  
				\toprule  
				&\multicolumn{4}{c}{\To\wikipedia}  &&\multicolumn{4}{c}{\To\cnn}   \\
				\cmidrule{2-5}
				\cmidrule{7-10}
				&\multicolumn{2}{c}{\wiki\To\wiki}  &\multicolumn{2}{c}{\cnn\To\wiki} &&\multicolumn{2}{c}{\cnn\To\cnn}  &\multicolumn{2}{c}{\wiki\To\cnn}  \\
				            &\Acc (\%)       &\Fscore (\%)  &\Acc (\%)     &\Fscore (\%)  && \Acc (\%)    &\Fscore (\%)   & \Acc (\%)    &\Fscore (\%)   \\ \midrule
				\majority   &57.3       &\z0.0   &57.3    & \z0.0   && 50.6    &  \z0.0  & 50.6     & \z0.0 \\
				\bow        &57.3       &\z0.0   &57.3    &\z0.0    && 50.6    &\z0.0    &50.6      & \z0.0 \\
				\bilstm     &56.2       &12.7    &57.3    &\z0.0    && 51.7    & 25.1    & 50.2     & \z3.0   \\ 
				\infersent  &57.3       &\z0.0   &57.3    &\z0.0    && 50.6    &\z0.0    & 50.6     &\z0.0  \\
				\skipthought&57.3       &\z0.0   &57.3    &\z0.0    && 50.6    &\z0.0    & 50.6     &\z0.0  \\
				\bertbase   &65.3       &35.7    &61.2    &21.1     && 80.8    & 71.6    & 57.0     &23.5  \\ 
				\bertlarge  &67.0       &39.6    &64.0&29.1 && 82.4    & 74.8    & 61.5     &35.9   \\ 
				\xlnetbase  &67.8       &45.0    &62.2    &22.4     && 91.2    & 86.6    & 64.0     &43.3   \\ 
				\xlnetlarge &72.9        &55.4   &62.8    &22.2     && \B{96.9}&95.0 &80.7 &73.8 \\
				\robertabase&69.5       &47.0    &63.2    &26.1     && 92.5    &88.8     & 77.6     &68.1 \\
				\robertalarge&76.1      &59.8    &63.7    &24.6     && 96.0    &94.5     & 88.3     &83.5  \\
				\albertlarge &70.7  &49.6   &63.8  &24.9  & &93.4  &90.8  &72.6   &61.5  \\ 
				\albertxxlarge &\B{81.7}  &\B{71.5}  &\B{66.6}  &\B{33.2}  &  &\B{96.9}   &\B{95.9}  &\B{89.1}  &\B{86.7}  \\ \midrule
				\albertxxlargefreeze &57.3 &0.0 &N/A &N/A & &50.6 &0.3 &N/A &N/A \\ \midrule
				Human & 66.6 & 35.9 &66.6 &35.9 &&\multicolumn{2}{c}{74.0}  &\multicolumn{2}{c}{57.8} \\ \bottomrule
			\end{tabular}
			
	\end{threeparttable}
	\caption{Experimental results over \wikipedia and \cnn, in both in-domain and cross-domain settings. \Acc is at the document level and \Fscore is at the sentence level.
	} 
	\label{experimental_results} 
\end{table*}

Our focus is on the task, dataset, and how existing models perform at document coherence prediction rather than modelling novelty, and we thus experiment with pre-existing pre-trained models. The models are as follows, each of which is fed into an MLP layer with a softmax output.

\paragraph{\bow:} Average the word embeddings for the combined document (sentence $+$ sequence of sentences in the document), based on pretrained 300D {GloVe} embeddings trained on a 840B-token corpus \cite{Pennington:14}.

\paragraph{\bilstm:} Feed the sequence of words in the combined document into a single-layer 512D \bilstm with average-pooling; word embeddings are initialised as with \bow. 

\paragraph{\infersent:} Generate representations for the sentence and document with \infersent \cite{Conneau:17}, and concatenate the two; \infersent is based on a Bi-LSTM with a max-pooling layer, trained on SNLI \cite{Bowman:15}.

\paragraph{\skipthought:} Generate representations for the sentence and document with \skipthought \cite{Kiros:2015}, and concatenate the two; \skipthought is an encoder--decoder model where the encoder extracts generic sentence embeddings and the decoder reconstructs surrounding sentences of the encoded sentence. 



\paragraph{\bert:} Generate representations for the concatenated sentence and document with \bert \cite{Devlin:19}, which was pretrained on the tasks of masked language modelling and next sentence prediction over \wikipedia and BooksCorpus \cite{Zhu:15}; we experiment with both \bertlarge and \bertbase (the cased versions).

\paragraph{\roberta:} Generate representations for the concatenated sentence and document with \roberta \cite{Liu:19}, which was pretrained on the task of masked language modelling (dynamically masking) and each input consisting of continuous sentences from the same document or multiple documents (providing broader context) over Cc-news, OpenWebTextCorpus, and STORIES \cite{Trinh:18}, in addition to the same data \bert was pretrained on; we experiment with both \robertalarge and \robertabase. 

\paragraph{\albert:} Generate representations for the concatenated sentence and document with \albert \cite{Lan:20}, which was pretrained over the same dataset as \bert but replaces the next sentence prediction objective with a sentence-order prediction objective, to model document coherence; we experiment with both \albertlarge and \albertxxlarge.

\paragraph{\xlnet:} Generate representations for the concatenated sentence and document with \xlnet \cite{Yang:19}, which was pretrained using a permutation language modelling objective over datasets including \wikipedia, BooksCorpus, Giga5 \cite{Parker:11}, ClueWeb 2012-B \cite{Callan:09}, and Common Crawl; we experiment with both \xlnetlarge and \xlnetbase (the cased versions). Although \xlnetlarge is used in removing data artefacts when selecting the intruder sentences, our experiments suggest that the comparative results across models (with or without artefact filtering) are robust.

\section{Experiments}

\subsection{Preliminary Results}
\label{general_performance}

In our first experiments, we train the various models across both \wikipedia and \cnn, and evaluate them in-domain and cross-domain. We are particularly interested in the cross-domain setting, to test the true ability of the model to detect document incoherence, as distinct from overfitting to domain-specific idiosyncrasies. It is also worth mentioning that \bert, \roberta, \albert, and \xlnet are pretrained on multi-sentence Wikipedia data, and have potentially memorised sentence pairs, making in-domain experiments problematic for \wikipedia in particular. Also of concern in applying models to the automatically-generated data is that it is entirely possible that an intruder sentence is undetectable to a human, because no incoherence results from the sentence substitution (bearing in mind that only 58\% of documents in \tabref{injected_type_analysis} contained information structure inconsistencies).

From \tabref{experimental_results}, we can see that the simpler models
(\bow, \bilstm, \infersent, and \skipthought) perform only at the level
of \majority at the document level, for both \wikipedia and \cnn. At the
sentence level (\Fscore), once again the models perform largely at the
level of \majority (\Fscore = 0.0), other than \bilstm in-domain
for \wikipedia and \cnn. 
In the final row of the table, we also see that humans are much better at detecting whether documents are incoherent (at the document level) than identifying the position of intruder sentences (at the sentence level), and that in general, human performance is low. This is likely the result of the fact that there are only 58\% of documents in \tabref{injected_type_analysis} containing information structure inconsistencies. We only conducted crowdsourcing over \wikipedia due to budget limitations and the fact that the \cnn documents are available online, making dataset hacks possible.\footnote{To have a general idea about the difficulty of the \cnn dataset, one of the authors annotated 100 documents (50 coherent and 50 incoherent documents), randomly sampled from the test set.}

Among the pretrained LMs, \albertxxlarge achieves the best performance over \wikipedia and \cnn, at both the document and sentence levels. Looking closer at the \wikipedia results, we find that \bertlarge achieves a higher precision than \xlnetlarge (71.0\% vs.\ 60.3\%), while \xlnetlarge achieves a higher recall (51.3\% vs.\ 27.4\%). \albertxxlarge achieves a precision higher than \bertlarge (79.7\%) and a recall higher than \xlnetlarge (64.9\%), leading to the overall best performance. Over \cnn, \albertxxlarge, \robertalarge, and \xlnetlarge achieve high precision and recall (roughly 93.0\% to 97\%).\footnote{The higher performance for all models/humans over the \cnn dataset indicates that it is easier for models/humans to identify the presence of intruder sentences. This is can be explained by the fact that a large proportion of documents include named entities, making it easier to detect the intruder sentences. In addition, the database used to retrieve candidate intruder sentences is smaller compared to that of \wikipedia.} The competitive results for \albertxxlarge over \wikipedia and \cnn result from the pretraining strategies, especially the sentence-order prediction loss capturing document coherence in isolation, different from next sentence prediction loss which conflates topic prediction and coherence prediction in a lower-difficulty single task. The performance gap for \albert, \roberta, and \xlnet between the base and large models are bigger than that of \bert, suggesting that they benefit from greater model capacity.\footnote{We also performed experiments where the models were allowed to predict the first sentence as the intruder sentence. As expected, model performance drops, \eg, \Fscore of \xlnetlarge drops from 55.4\% to 47.9\%, reflecting both the increased complexity of the task and the lack of (at least) one previous sentence to provide document context.}

We also examine how pretrained LMs perform with only the classifier parameters being updated during training. Here, we focus on exclusively on \albertxxlarge, given its superiority. As shown in \figref{experimental_results}, the pretrained LM \albertxxlarge is unable to different coherent documents from incoherent ones, resulting into random guess, although it considers document coherence during pretraining. This indicates the necessity of finetuning LMs for document coherent understanding.  

Looking to the cross-domain results, again, \albertxxlarge achieves the best performance over both \wikipedia and \cnn. The lower results for \robertalarge and \xlnetlarge over \wikipedia may be due to both \roberta and \xlnet being pretrained over newswire documents, and fine-tuning over \cnn reducing the capacity of the model to generalise. \albert and \bert do not suffer from this as they are not pretrained over newswire documents. 
The substantial drop between the in- and cross-domain settings for \albert, \roberta, \xlnet, and \bert indicates that the models have limited capacity to learn a generalised representation of document coherence, in addition to the style differences between \wikipedia and \cnn.

\subsection{Results over the Existing Dataset}
\label{results_over_existing_dataset}

\begin{table}[t]
	\small
	\centering
	\scalebox{0.9}{ 
		\begin{threeparttable}
			\begin{tabular}{lcc}  
				\toprule  
				&Wiki\To Wiki & Ubuntu\To Wiki \\ \midrule
				\majority  &50.0  & 50.0 \\
                \albertxxlarge &96.8  &53.1  \\
                Human &98.0  & 98.0 \\  \midrule
              	&Ubuntu\To Ubuntu & Wiki\To Ubuntu \\ \midrule
                \majority  &50.0  & 50.0 \\
                \albertxxlarge &58.1  &58.7  \\
                Human &74.0  & 74.0 \\ 
                \bottomrule
			\end{tabular}
			
	\end{threeparttable}}
	\caption{\Acc for the dataset of \citet{Chen:19}.} 
	\label{result_over_existing} 

\end{table}

We also examine how \albertxxlarge performs over the coarse-grained dataset of \citet{Chen:19}, where 50 documents from each domain were annotated by a native English speaker. Performance is measured at the document level only, as the dataset does not include indication of which sentence is the intruder sentence. As shown in \tabref{result_over_existing}, \albertxxlarge achieves an \Acc of 96.8\% over the \wikipedia subset, demonstrating that our \wikipedia dataset is more challenging (\Acc of 81.7\%) and also underlining the utility of adversarial filtering in dataset construction. Given the considerably lower results, one could conclude that Ubuntu is a good source for a dataset. However, when one of the authors attempted to perform the task manually, they found the document-level task to be extremely difficult as it relied heavily on expert knowledge of Ubuntu packages, much more so than document coherence understanding.

In the cross-domain setting, there is a substantial drop over the \wikipedia dataset, which can be explained by \albertlarge failing to generate a representation of document coherence from the Ubuntu dataset, due to the high dependence on domain knowledge as described above, resulting in near-random results. The cross-domain results for \albertxxlarge over Ubuntu are actually marginally higher than the in-domain results but still close to random, suggesting that the in-domain model isn't able to capture either document coherence or domain knowledge, and underlining the relatively minor role of coherence for the Ubuntu dataset.

\subsection{Performance on Documents of Different Difficulty Levels}
\label{analysis_over_difficulty}

One concern with our preliminary experiments was whether the intruder sentences generate genuine incoherence in the information structure of the documents. We investigate this question by breaking down the results over the best-performing model (\albertxxlarge) based on the level of agreement between the human annotations and the generated gold-standard, for \wikipedia. The results are in \figref{human_models_accuracy_f1}, where the $x$-axis denotes the number of annotators who agree with the gold-standard: for example, ``2'' indicates that 2/5 annotators were able to assign the gold-standard labels to the documents. 

Our assumption is that the incoherent documents which humans fail to detect are actually not perceptibly incoherent,\footnote{Although the intruder sentence may lead to factual errors, the annotators were instructed not to do fact checking.} and that any advantage for the models over humans for documents with low-agreement (with respect to the gold-standard) is actually due to dataset artefacts. At the document level (\Acc), there is reasonable correlation between model and human performance (\ie, the model struggles on the same documents as the humans). 
At the sentence level (\Fscore), there is less discernible difference in model performance over documents of varying human difficulty. 

\begin{figure}[t!]
	\begin{center}
		\scalebox{1}{
			\definecolor{mycolor1}{rgb}{0.00000,0.44700,0.74100}%
\definecolor{mycolor2}{rgb}{0.85000,0.02500,0.09800}%
\begin{tikzpicture}

\begin{axis}[%
width=0.3\columnwidth,
scale only axis,
xmin=0,
xmax=5,
xlabel style={font=\color{white!15!black}},
xtick=data,
ymin=20,
ymax=100,
ylabel style={at={(1.1,1.15)},rotate=-90,},
ylabel={\Acc (\%)},
axis background/.style={fill=white},
legend style={at={(0.0,-0.3)}, anchor=north west, legend cell align=center, legend columns=2, align=center, draw=white!15!black},
]
\addplot [color=mycolor1, mark=o, mark options={solid, mycolor1}]
  table[row sep=crcr]{%
0	54.4\\
1	65.8\\
2	79.0\\
3	88.0\\
4	92.1\\
5	95.0\\
};
\addlegendentry{\albertxxlarge}

\addplot [color=mycolor2, mark=asterisk, mark options={solid, mycolor2}]
  table[row sep=crcr]{%
0	22.0\\
1	30.3\\
2	40.8\\
3	62.1\\
4	80.5\\
5	100.0\\
};
\addlegendentry{Humans}

\end{axis}

\begin{axis}[%
xshift=3.5cm,
width=0.3\columnwidth,
scale only axis,
xmin=0,
xmax=5,
xlabel style={font=\color{white!15!black}},
xtick=data,
ymin=0,
ymax=100,
ylabel style={at={(1.05,1.15)},rotate=-90,},
ylabel={\Fscore (\%)},
axis background/.style={fill=white},
]
\addplot [color=mycolor1, mark=o, mark options={solid, mycolor1}]
table[row sep=crcr]{%
	0	61.6\\
	1	73.0\\
	2	80.8\\
	3	76.7\\
	4	75.0\\
	5	80.7\\
};

\addplot [color=mycolor2, mark=asterisk, mark options={solid, mycolor2}]
table[row sep=crcr]{%
	0	0.0\\
	1	29.0\\
	2	49.3\\
	3	42.0\\
	4	52.1\\
	5	100.0\\
};

\end{axis}
\end{tikzpicture}%
		}
	\end{center}
	\caption{\albertxxlarge vs.\ humans.}
	\label{human_models_accuracy_f1}
\end{figure}

\subsection{Analysis over Documents with High Human Agreement}
\begin{table}[t]
  \small
	\centering
		\begin{threeparttable}
			\begin{tabular}{lcc@{}r}  
				\toprule  

				Humans &\# $-$intruder docs & \#  $+$intruder docs \\  \midrule
				Coherent 	& 1385       &177           \\ 
				Incoherent	& 11   & 404       \\ \bottomrule
			\end{tabular}	
	\end{threeparttable}
	\caption{Statistics over documents where all 5 humans agree, where $-$intruder/$+$intruder indicates the documents without/with an intruder sentence.} 
	\label{correct_vs_wrong} 
\end{table}

To understand the relationship between human-assigned labels and the gold-standard, we further examine documents where all 5 annotators agree, noting that human-assigned labels can potentially be different from the gold-standard here. \tabref{correct_vs_wrong} shows the statistics of humans over these documents, with regard to whether there is an intruder sentence in the documents. Encouragingly, we can see that humans tend to agree more over coherent documents (documents without any intruder sentences) than incoherent documents (documents with an intruder sentence). Examining the 11 original coherent documents which were annotated as incoherent by all annotators, we find out that there is a break in information flow due to references or urls, even though there is no intruder sentence. For documents with an intruder sentence ($+$ intruder), where humans disagree with the gold-standard (humans perceive the documents as coherent or the position of the intruder sentence to be other than the actual intruder sentence), we find that 98\% of the documents are considered to be coherent. We randomly sampled 100 documents from these documents and examined whether the intruder sentence results in a break in information flow. We find that fact-checking is needed to identify the intruder sentence for 93\% of the documents.\footnote{Here, the high percentage of incoherent documents with factual inconsistencies does not necessarily point to a high percentage of factual inconsistency in the overall dataset, as humans are more likely to agree with the gold-standard for coherent documents.}

\begin{table}[t]
	\centering
	\scalebox{0.78}{ 
	\begin{threeparttable}
		\begin{tabular}{lcccc}  
			\toprule  
			\multirow{2}{*}{} 
			&\multicolumn{2}{c}{\wiki\To\wiki}  &\multicolumn{2}{c}{\cnn\To\wiki}  \\
			&\Acc (\%)        &\Fscore (\%)  &\Acc (\%)  &\Fscore (\%)  \\ \midrule
			\majority &70.6   &\z0.0    &70.6   &\z0.0  \\
			\bertlarge &76.7   &42.0    &75.4   & 36.9  \\
			\xlnetlarge &79.1   &57.0   &76.6   & 35.4   \\
			\robertalarge &82.0 &59.6   &77.3   &37.4\\ 
			\albertxxlarge &\B{85.9}  &\B{68.8}  &\B{78.8} &\B{42.9}  \\\midrule
			Human &79.5   &45.4  &79.5   &45.4  \\ \bottomrule
		\end{tabular}
		
  	\end{threeparttable}}
	\caption{Results over documents annotated consistently by all 5 annotators, where annotations can be the same as or different from gold-standard.} 
	\label{result_over_5} 
\end{table}

\tabref{result_over_5} shows the performance over the \wikipedia documents that are annotated consistently by all 5 annotators (from \tabref{correct_vs_wrong}). Consistent with the results from \tabref{experimental_results}, \albertxxlarge achieves the best performance both in- and cross-domain. To understand the different behaviours of humans and \albertxxlarge, we analyse documents which only humans got correct, only \albertxxlarge got correct, or neither humans nor \albertxxlarge got correct, as follows:

\begin{compactenum}
	\item Humans only: 7 incoherent ($+$intruder) and 73 coherent ($-$intruder) documents
	\item \albertxxlarge only: 181 incoherent ($+$intruder) (of which we found 97\% to require fact-checking\footnote{There are 4 documents that humans identify as incoherent based on the wrong intruder sentence, due to the intruder sentence leading to a misleading factual inconsistency.}) and 9 coherent ($-$intruder) documents (of which 8 contain urls/references, which confused humans)
	\item Neither humans nor models: 223 incoherent ($+$intruder) (of which 98.2\% and 77.1\% were predicted to be coherent by humans and \albertxxlarge, respectively, and for the remainder, the wrong intruder sentence was identified) and 2 coherent ($-$intruder) documents (both of which were poorly organised, confusing allcomers)
\end{compactenum} 

Looking over the incoherent documents which require fact-checking, no obvious differences are discernible between the documents that \albertxxlarge predicts correctly and those it misses. Our assumption here is that \albertxxlarge is biased by the pretraining dataset, and that many of the cases where it makes the wrong prediction are attributable to mismatches between the text in our dataset and the Wikipedia version used in pretraining the model.


\subsection{Question Revisited}
\label{question_revisited}


\B{Q1: Do models truly capture the intrinsic properties of document coherence?}

\B{A:} It is certainly true that models which incorporate a more explicit notion of document coherence into pretraining (e.g.\ \albert) tend to perform better. In addition, larger-context models (\roberta) and robust training strategies (\xlnet) during pretraining are also beneficial for document coherent understanding. This suggests a tentative yes, but there were equally instances of strong disagreement with human intuitions and model predictions for the better-performing models and evidence to suggest that the models were performing fact-checking at the same time as coherence modelling.   

\B{Q2: What types of document incoherence can/can't these
models detect?}

\B{A:} Over incoherent documents resulting from fact inconsistencies, where humans tend to fail, the better-performing models can often make correct predictions; over incoherent documents with information structure or logical inconsistencies which humans can easily detect, \albertlarge, \robertalarge, and \xlnetlarge achieve an \Acc $\ge 87$\%, showing that they can certainly capture information structure and logical inconsistencies to a high degree. That said, the fact that they misclassify clearly coherent documents as incoherent suggests that are in part lacking in their ability to capture document coherence. We thus can conclude that they can reliably identify intruder sentences which result in a break in information structure or logical flow, but are imperfect models of document coherence. 

\section{Linguistic Probes}
\label{language_probes}

To further examine the models, we constructed a language probe dataset. 

\begin{table*}[t] 
	\small
	\centering
	\scalebox{0.95}{ 
		\begin{threeparttable}
			\begin{tabular}{lccccccccccc@{}r}  
                          \toprule  
                          &\multicolumn{2}{C{2.5cm}}{\gender}  &&\multicolumn{2}{C{2.5cm}}{\animacy$\downarrow$} &&\multicolumn{2}{C{2.5cm}}{\animacy$\uparrow$}  &&\multicolumn{2}{C{2.5cm}}{\past}  \\
                          \cmidrule{2-3}
                          \cmidrule{5-6}
                          \cmidrule{8-9}
                          \cmidrule{11-12}

                                                   &\Fscore&$\Delta$\Fscore   &&\Fscore&$\Delta$\Fscore &&\Fscore&$\Delta$\Fscore  &&\Fscore&$\Delta$\Fscore\\ \midrule
                          \bertlarge               &26.5   & $\deltapos$65.3   &&26.3   & $\deltapos$53.2 &&33.6   & $\deltapos$45.1  &&35.6   &$\deltapos$42.1 \\ 
                          \xlnetlarge	           &55.8   &$\deltapos$41.6    &&50.0   &$\deltapos$45.2  &&64.0   &$\deltapos$23.5   &&64.9   &$\deltapos$16.9   \\ 
                          \robertalarge            &64.9   &$\deltapos$32.5    &&50.7   &$\deltapos$38.3  &&59.7   &$\deltapos$21.7   &&69.2   &$\deltapos$19.9 \\ 
                          \albertxxlarge           &74.0   &$\deltapos$25.4    &&71.8   &\z$\deltapos$8.5 &&81.0   &\z$\deltapos$2.9  &&79.8   &\z$\deltapos$4.3  \\[1ex] 
                          \bertlarge[\CNNtrained]    &23.9   &$\deltapos$70.0    &&22.2   &$\deltapos$60.2  &&27.6   &$\deltapos$51.4   &&30.6   &$\deltapos$14.7\\  
                          \xlnetlarge[\CNNtrained]   &13.6   &$\deltapos$83.1    &&10.0   &$\deltapos$71.3  &&\z8.0  &$\deltapos$71.8   &&23.2   &$\deltapos$27.6 \\ 
                          \robertalarge[\CNNtrained] &15.4   &$\deltapos$82.4    &&\z7.9  &$\deltapos$64.4  &&\z9.8  &$\deltapos$73.3   &&23.4   &$\deltapos$40.0  \\ 
                          \albertxxlarge[\CNNtrained]&21.6   &$\deltapos$72.8    &&20.2   &$\deltapos$51.8  &&27.6   &$\deltapos$33.4   &&38.0   &$\deltapos$30.4  \\ \midrule
                          Human	                   &35.8   &$\deltapos$53.4    &&36.6   &$\deltapos$45.3  &&29.8   &$\deltapos$53.9   &&40.9   &$\deltapos$34.4 \\ \bottomrule
                          \\

                          \toprule
                                                        &\multicolumn{2}{C{2.5cm}}{\conjunction}  &&\multicolumn{2}{C{2.5cm}}{\demonstrative} &&\multicolumn{2}{C{2.5cm}}{\negation}  &&\multicolumn{2}{C{2.5cm}}{\numbermanipulation}  \\
                          \cmidrule{2-3}
                          \cmidrule{5-6}
                          \cmidrule{8-9}
                          \cmidrule{11-12}
                                                   &\Fscore&$\Delta$\Fscore  &&\Fscore&$\Delta$\Fscore  &&\Fscore&$\Delta$\Fscore  &&\Fscore&$\Delta$\Fscore\\ \midrule
                          \bertlarge	           &51.9   &$\deltapos$17.3   &&34.8   &$\deltapos$15.6   &&34.5   &$\deltapos$32.2   &&32.5   &$\deltapos$31.2  \\ 
                          \xlnetlarge	           & 68.6  &\z$\deltapos$3.6  &&55.4   &\z\z0.0          &&57.7   &\z$\deltapos$8.9  &&50.7   &$\deltapos$11.3 \\   
                          \robertalarge            &73.0   &\z$\deltapos$0.7  &&57.9   &\z\z0.0          &&68.4   &$\deltapos$10.9   &&54.2   &$\deltapos$20.0  \\
                          \albertxxlarge           &83.5   &\z$\deltaneg$1.6&&75.2   &\z$\deltapos$1.3  &&79.5   &\z$\deltapos$2.9  &&63.9   &$\deltapos$10.4 \\[1ex]
                          \bertlarge[\CNNtrained]	   &38.2   &\z$\deltaneg$1.4&&35.6   &\z$\deltaneg$5.7&&28.8   &\z$\deltapos$4.2  &&19.6   &$\deltapos$11.7  \\  
                          \xlnetlarge[\CNNtrained]   &31.0   &\z\z0.0          &&14.1   &\z\z0.0          &&15.7   &$\deltapos$11.8   &&15.2   &$\deltapos$13.1 \\
                          \robertalarge[\CNNtrained] &33.9   &\z$\deltapos$1.4  &&17.8   &\z\z0.0          &&21.0   &$\deltapos$12.4   &&18.3   &$\deltapos$23.6 \\
                          \albertxxlarge[\CNNtrained]&41.6   &\z$\deltapos$1.3  &&30.9   &\z\z0.0          &&28.1   &$\deltapos$19.2   &&23.0   &$\deltapos$16.0  \\ \midrule                                           
                          Human	                   &40.5   &\z$\deltapos$8.7  &&38.0   &\z$\deltapos$1.0  &&40.4   &$\deltapos$36.8   &&37.3   &$\deltapos$24.2 \\ \bottomrule
			\end{tabular}	
	\end{threeparttable}}
	\caption{Results over language probes in incoherent \wikipedia test documents. \bertlarge[\CNNtrained], \xlnetlarge[\CNNtrained], \robertalarge[\CNNtrained], and \albertxxlarge[\CNNtrained] are trained over \cnn, while \bertlarge, \xlnetlarge, \robertalarge, and \albertxxlarge are trained over \wikipedia. Here, \Fscore is over the original incoherent documents (excluding linguistic probes), and $\Delta$\Fscore indicates the absolute performance difference resulting from incorporating linguistic probes.}
	\label{injected_language_probe_relative_result} 
\end{table*}

\subsection{Linguistic Probe Dataset Construction}

We handcrafted adversarial instances based on a range of linguistic phenomena which generate information structure inconsistencies. In constructing such a dataset, minimal modifications were made to the original sentences, to isolate the effect of the linguistic probe. For each phenomenon, we hand-constructed roughly 100 adversarial instances by modifying intruder sentences in \textit{incoherent} \wikipedia test documents which were manually pre-filtered for ease of detection/lack of confounding effects in the original text. That is, the linguistic probes for the different phenomena were manually added to incoherent test documents, within intruder sentences; our interest here is whether the addition of the linguistic probes makes it easier for the models to detect the incoherence. Importantly, we do not provide any additional training data, meaning there is no supervision signal specific to the phenomena.  There are roughly 8$\times$100 instances in total,\footnote{There are 100 instances for each phenomenon except for \demonstrative, where there were only 95 instances in the \wikipedia test data with singular demonstratives.}  with the eight phenomena being: 
\begin{compactenum}
	\item gender pronoun flip (\gender), converting a pronoun to its opposite gender, \eg \example{she}$\,\to\,$\example{he};
	\item animacy downgrade (\animacy$\downarrow$), downgrading pronouns and possessive determiners to their inanimate versions, \eg \example{she}/\example{he}/\example{her}/\example{him}$\,\to\,$\example{it}, and \example{her}/\example{his}$\,\to\,$\example{its};
	\item animacy upgrade (\animacy$\uparrow$), upgrading pronouns and possessive determiners to their third person version, \eg \example{it}$\,\to\,$\example{she}/\example{he}/\example{her}/\example{him}, and \example{its}$\,\to\,$\example{her}/\example{his};
	\item singular demonstrative flip (\demonstrative), converting singular demonstratives to plural ones, \eg \example{this}$\,\to\,$\example{these} and \example{that}$\,\to\,$\example{those};
	\item conjunction flip (\conjunction), converting conjunctions to their opposites, \eg \example{but}$\,\to\,$\example{and therefore}, \example{and}$\,\to\,$\example{but}, \example{although}$\,\to\,$\example{therefore}, and vice versa;
	\item past tense flip (\past), converting past to future tense, \eg \example{was}$\,\to\,$\example{will be} and \example{led}$\,\to\,$\example{will lead};
	\item sentence negation (\negation), negating the sentence, \eg \example{He has [a] ... warrant ...}$\,\to\,$\example{He doesn't have [a] ... warrant ...};
	\item number manipulation (\numbermanipulation), changing numbers to implausible values, \eg \example{He served as Chief Operating Officer ... from 2002 to 2005}$\,\to\,$\example{He served as Chief Operating Officer ... from 200 BCE to 201 BCE} and \example{Line 11 has a length of 51.7 km and a total of 18 stations.}$\,\to\,$\example{Line 11 has a length of 51.7 m and a total of 1.8 stations.}.
\end{compactenum}

All the probes generate syntactically correct sentences, and the first four generally lead to sentences which are also semantically felicitous, with the incoherence being at the document level. For example, in \example{He was never convicted and was out on parole within a few years}, if we replace \example{he} with \example{she}, the sentence is felicitous, but if the focus entity in the preceding and subsequent sentences is a male, the information flow will be disrupted. 

The last four language probes are crafted to explore the capacity of a model to capture commonsense reasoning, in terms of discourse relationships, tense and polarity awareness, and understanding of numbers. For \conjunction, we only focus on explicit connectives within a sentence. For \past, there can be intra-sentence inconsistency if there are time-specific signals, failing which broader document context is needed to pick up on the tense flip. Similarly for \negation and \numbermanipulation, the change can lead to inconsistency either intra- or inter-sententially. For example, \example{He did not appear in more than 400 films between 1914 and 1941 ...} is intra-sententially incoherent.

\subsection{Experimental Results}

\tabref{injected_language_probe_relative_result} lists the performance of pretrained LMs at recognising intruder sentences within incoherent documents, with and without the addition of the respective linguistic probes.\footnote{Results for coherent documents are omitted due to space.}
For a given model, we break down the results across probes into two columns: the first column (``\Fscore'') shows the sentence-level performance over the original intruder sentence (without the addition of the linguistic probe), and the second column (``$\Delta$\Fscore'') shows the absolute difference in performance with the addition of the linguistic probe. Our expectation is that results should improve on average with the inclusion of the linguistic probe (i.e.\ $\Delta$\Fscore values should be positive), given that we have reinforced the incoherence generated by the intruder sentence.

All models achieve near-perfect results with \gender linguistic probes (i.e.\ the sum of \Fscore and $\Delta$\Fscore is close to 100), and are also highly successful at detecting \animacy~mismatches and \past (the top half of \tabref{injected_language_probe_relative_result}). For the probes in the bottom half of the table, none of the three models except \albertxxlarge performs particularly well, especially for \demonstrative. For each linguistic probe, we observe that the pretrained LMs can more easily detect incoherent text with the addition of these lexical/grammatical inconsistencies (except for \xlnetlarge and \albertxxlarge over \demonstrative and \albertxxlarge over \conjunction).

In the cross-domain setting, the overall performance of \xlnetlarge[\CNNtrained] and \albertxxlarge[\CNNtrained] drops across all linguistic probes, but the absolute gain through the inclusion of the linguistic probe is almost universally larger, suggest that while domain differences hurt the models, they are attuned to the impact of linguistic probes on document coherence and thus learning some more general properties of document (in)coherence. On the other hand, \bertlarge[\CNNtrained] (over \gender, \animacy$\downarrow$, and \animacy$\uparrow$) and \robertalarge[\CNNtrained] (\gender and \animacy$\uparrow$) actually perform better than in-domain. \robertalarge[\CNNtrained] achieves the best overall performance over \gender, \animacy$\uparrow$, and \numbermanipulation while \albertxxlarge[\CNNtrained] achieves the best overall performance over \past, \conjunction, \demonstrative, and \negation. The reason that the models tend to struggle with \demonstrative and \conjunction is not immediately clear, and will be explored in future work.

We also conducted human evaluations on this dataset via Amazon Mechanical Turk, based on the same methodology as described in \secref{sec:verification} (without explicit instruction to look out for linguistic artefacts, and with a mixture of coherent and incoherent documents, as per the original annotation task). As detailed in \tabref{injected_language_probe_relative_result}, humans generally benefit from the inclusion of the linguistic probes. Largely consistent with the results for the models, humans are highly sensitised to the effects of \gender, \animacy, \past, and \negation, but largely oblivious to the effects of \demonstrative and \conjunction. Remarkably, the best models (\albertxxlarge and \robertalarge) perform on par with humans in the in-domain setting, but are generally well below humans in the cross-domain setting.

\section{Conclusion}
\label{conclusion}

We propose the new task of detecting whether there is an intruder sentence in a document, generated by replacing an original sentence with a similar sentence from a second document. To benchmark model performance over this task, we construct a large-scale dataset consisting of documents from English \wikipedia and \cnn news articles. Experimental results show that pretrained LMs which incorporate larger document contexts in pretraining perform remarkably well in-domain, but experience a substantial drop cross-domain. In follow-up analysis based on human annotations, substantial divergences from human intuitions were observed, pointing to limitations in their ability to model document coherence. Further results over a linguistic probe dataset show that pretrained models fail to identify some linguistic characteristics that affect document coherence, suggesting room to improve for them to truly capture document coherence, and motivating the construction of a dataset with intruder text at the intra-sentential level. 

\bibliographystyle{acl_natbib}
\bibliography{main-2775-Shen}

\begin{thebibliography}{85}
\expandafter\ifx\csname natexlab\endcsname\relax\def\natexlab#1{#1}\fi

\bibitem[{Adouane et~al.(2018{\natexlab{a}})Adouane, Bernardy, and
  Dobnik}]{Adouane:18a}
Wafia Adouane, Jean-Philippe Bernardy, and Simon Dobnik. 2018{\natexlab{a}}.
\newblock Improving neural network performance by injecting background
  knowledge: {D}etecting code-switching and borrowing in {A}lgerian texts.
\newblock In \emph{Proceedings of the Third Workshop on Computational
  Approaches to Linguistic Code-Switching}, pages 20--28.

\bibitem[{Adouane et~al.(2018{\natexlab{b}})Adouane, Dobnik, Bernardy, and
  Semmar}]{Adouane:18b}
Wafia Adouane, Simon Dobnik, Jean-Philippe Bernardy, and Nasredine Semmar.
  2018{\natexlab{b}}.
\newblock A comparison of character neural language model and bootstrapping for
  language identification in multilingual noisy texts.
\newblock In \emph{Proceedings of the Second Workshop on Subword/Character
  {LE}vel Models}, pages 22--31.

\bibitem[{Alikhani et~al.(2019)Alikhani, Nag~Chowdhury, de~Melo, and
  Stone}]{Alikhani:19}
Malihe Alikhani, Sreyasi Nag~Chowdhury, Gerard de~Melo, and Matthew Stone.
  2019.
\newblock {CITE}: {A} corpus of image-text discourse relations.
\newblock In \emph{Proceedings of the 2019 Conference of the North {A}merican
  Chapter of the Association for Computational Linguistics: Human Language
  Technologies, Volume 1 (Long and Short Papers)}, pages 570--575.

\bibitem[{Asher and Lascarides(2003)}]{Asher:03}
Nicholas~M. Asher and Alex Lascarides. 2003.
\newblock \emph{Logics of Conversation}.
\newblock Studies in Natural Language Processing. Cambridge University Press.

\bibitem[{Barzilay and Lapata(2005)}]{Barzilay:05}
Regina Barzilay and Mirella Lapata. 2005.
\newblock Modeling local coherence: {A}n entity-based approach.
\newblock In \emph{Proceedings of the 43rd Annual Meeting of the Association
  for Computational Linguistics}, pages 141--148.

\bibitem[{Barzilay and Lapata(2008)}]{Barzilay:08}
Regina Barzilay and Mirella Lapata. 2008.
\newblock Modeling local coherence: {A}n entity-based approach.
\newblock \emph{Computational Linguistics}, 34(1):1--34.

\bibitem[{Belinkov et~al.(2017)Belinkov, Durrani, Dalvi, Sajjad, and
  Glass}]{Belinkov:17}
Yonatan Belinkov, Nadir Durrani, Fahim Dalvi, Hassan Sajjad, and James Glass.
  2017.
\newblock What do neural machine translation models learn about morphology?
\newblock In \emph{Proceedings of the 55th Annual Meeting of the Association
  for Computational Linguistics (Volume 1: Long Papers)}, pages 861--872.

\bibitem[{Belinkov and Glass(2019)}]{Belinkov:19}
Yonatan Belinkov and James Glass. 2019.
\newblock Analysis methods in neural language processing: {A} survey.
\newblock \emph{Transactions of the Association for Computational Linguistics},
  7:49--72.

\bibitem[{Blevins et~al.(2018)Blevins, Levy, and Zettlemoyer}]{Blevins:18}
Terra Blevins, Omer Levy, and Luke Zettlemoyer. 2018.
\newblock Deep {RNN}s encode soft hierarchical syntax.
\newblock In \emph{Proceedings of the 56th Annual Meeting of the Association
  for Computational Linguistics (Volume 2: Short Papers)}, pages 14--19.

\bibitem[{Bowman et~al.(2015)Bowman, Angeli, Potts, and Manning}]{Bowman:15}
Samuel~R. Bowman, Gabor Angeli, Christopher Potts, and Christopher~D. Manning.
  2015.
\newblock A large annotated corpus for learning natural language inference.
\newblock In \emph{Proceedings of the 2015 Conference on Empirical Methods in
  Natural Language Processing}, pages 632--642.

\bibitem[{Callan et~al.(2009)Callan, Hoy, Yoo, and Zhao}]{Callan:09}
Jamie Callan, Mark Hoy, Changkuk Yoo, and Le~Zhao. 2009.
\newblock Clueweb09 data set.
\newblock \url{https://lemurproject.org/clueweb09/} Accessed: 15.12.2019.

\bibitem[{Chandu et~al.(2019)Chandu, Nyberg, and Black}]{Chandu:19}
Khyathi Chandu, Eric Nyberg, and Alan~W Black. 2019.
\newblock Storyboarding of recipes: Grounded contextual generation.
\newblock In \emph{Proceedings of the 57th Annual Meeting of the Association
  for Computational Linguistics}, pages 6040--6046.

\bibitem[{Chen et~al.(2017)Chen, Fisch, Weston, and Bordes}]{Chen:17}
Danqi Chen, Adam Fisch, Jason Weston, and Antoine Bordes. 2017.
\newblock Reading {W}ikipedia to answer open-domain questions.
\newblock In \emph{Proceedings of the 55th Annual Meeting of the Association
  for Computational Linguistics (Volume 1: Long Papers)}, pages 1870--1879.

\bibitem[{Chen et~al.(2019)Chen, Chu, and Gimpel}]{Chen:19}
Mingda Chen, Zewei Chu, and Kevin Gimpel. 2019.
\newblock Evaluation benchmarks and learning criteria for discourse-aware
  sentence representations.
\newblock In \emph{Proceedings of the 2019 Conference on Empirical Methods in
  Natural Language Processing and the 9th International Joint Conference on
  Natural Language Processing}, pages 649--662.

\bibitem[{Cheng et~al.(2020)Cheng, Yi, Chen, Zhang, and Hsieh}]{Cheng:18}
Minhao Cheng, Jinfeng Yi, Pin{-}Yu Chen, Huan Zhang, and Cho{-}Jui Hsieh. 2020.
\newblock Seq2sick: Evaluating the robustness of sequence-to-sequence models
  with adversarial examples.
\newblock In \emph{Proceedings of the Thirty-Fourth AAAI Conference on
  Artificial Intelligence}, pages 3601--3608.

\bibitem[{Clercq et~al.(2014)Clercq, Hoste, Desmet, van Oosten, Cock, and
  Macken}]{Clercq:14}
Orph{\'{e}}e~De Clercq, V{\'{e}}ronique Hoste, Bart Desmet, Philip van Oosten,
  Martine~De Cock, and Lieve Macken. 2014.
\newblock Using the crowd for readability prediction.
\newblock \emph{Natural Language Engineering}, 20(3):293--325.

\bibitem[{Conneau et~al.(2017)Conneau, Kiela, Schwenk, Barrault, and
  Bordes}]{Conneau:17}
Alexis Conneau, Douwe Kiela, Holger Schwenk, Lo{\"\i}c Barrault, and Antoine
  Bordes. 2017.
\newblock Supervised learning of universal sentence representations from
  natural language inference data.
\newblock In \emph{Proceedings of the 2017 Conference on Empirical Methods in
  Natural Language Processing}, pages 670--680.

\bibitem[{Conneau et~al.(2018)Conneau, Kruszewski, Lample, Barrault, and
  Baroni}]{Baroni:18}
Alexis Conneau, German Kruszewski, Guillaume Lample, Lo{\"\i}c Barrault, and
  Marco Baroni. 2018.
\newblock What you can cram into a single {\$}{\&}!{\#}* vector: {P}robing
  sentence embeddings for linguistic properties.
\newblock In \emph{Proceedings of the 56th Annual Meeting of the Association
  for Computational Linguistics (Volume 1: Long Papers)}, pages 2126--2136.

\bibitem[{Devlin et~al.(2019)Devlin, Chang, Lee, and Toutanova}]{Devlin:19}
Jacob Devlin, Ming-Wei Chang, Kenton Lee, and Kristina Toutanova. 2019.
\newblock {BERT}: {P}re-training of deep bidirectional transformers for
  language understanding.
\newblock In \emph{Proceedings of the 2019 Conference of the North {A}merican
  Chapter of the Association for Computational Linguistics: Human Language
  Technologies, Volume 1 (Long and Short Papers)}, pages 4171--4186.

\bibitem[{Elsner et~al.(2007)Elsner, Austerweil, and Charniak}]{Elsner:07}
Micha Elsner, Joseph Austerweil, and Eugene Charniak. 2007.
\newblock A unified local and global model for discourse coherence.
\newblock In \emph{Human Language Technologies 2007: The Conference of the
  North {A}merican Chapter of the Association for Computational Linguistics;
  Proceedings of the Main Conference}, pages 436--443.

\bibitem[{Elsner and Charniak(2011)}]{Elsner:11}
Micha Elsner and Eugene Charniak. 2011.
\newblock Extending the entity grid with entity-specific features.
\newblock In \emph{Proceedings of the 49th Annual Meeting of the Association
  for Computational Linguistics: Human Language Technologies}, pages 125--129.

\bibitem[{Fan et~al.(2019)Fan, Lewis, and Dauphin}]{Fan:19}
Angela Fan, Mike Lewis, and Yann Dauphin. 2019.
\newblock Strategies for structuring story generation.
\newblock In \emph{Proceedings of the 57th Annual Meeting of the Association
  for Computational Linguistics}, pages 2650--2660.

\bibitem[{Grice(2002)}]{Grice:75}
Herbert~Paul Grice. 2002.
\newblock Logic and conversation.
\newblock \emph{Foundations of Cognitive Psychology}, pages 719--732.

\bibitem[{Guinaudeau and Strube(2013)}]{Guinaudeau:13}
Camille Guinaudeau and Michael Strube. 2013.
\newblock Graph-based local coherence modeling.
\newblock In \emph{Proceedings of the 51st Annual Meeting of the Association
  for Computational Linguistics (Volume 1: Long Papers)}, pages 93--103.

\bibitem[{Gulordava et~al.(2018)Gulordava, Bojanowski, Grave, Linzen, and
  Baroni}]{Gulordava:18}
Kristina Gulordava, Piotr Bojanowski, Edouard Grave, Tal Linzen, and Marco
  Baroni. 2018.
\newblock Colorless green recurrent networks dream hierarchically.
\newblock In \emph{Proceedings of the 2018 Conference of the North {A}merican
  Chapter of the Association for Computational Linguistics: Human Language
  Technologies, Volume 1 (Long Papers)}, pages 1195--1205.

\bibitem[{Hermann et~al.(2015)Hermann, Ko\v{c}isk\'{y}, Grefenstette, Espeholt,
  Kay, Suleyman, and Blunsom}]{Hermann:15}
Karl~Moritz Hermann, Tom\'{a}\v{s} Ko\v{c}isk\'{y}, Edward Grefenstette, Lasse
  Espeholt, Will Kay, Mustafa Suleyman, and Phil Blunsom. 2015.
\newblock Teaching machines to read and comprehend.
\newblock In \emph{Proceedings of the 28th International Conference on Neural
  Information Processing Systems - Volume 1}, pages 1693--1701.

\bibitem[{Hewitt and Liang(2019)}]{Hewitt:19a}
John Hewitt and Percy Liang. 2019.
\newblock Designing and interpreting probes with control tasks.
\newblock In \emph{Proceedings of the 2019 Conference on Empirical Methods in
  Natural Language Processing and the 9th International Joint Conference on
  Natural Language Processing}, pages 2733--2743.

\bibitem[{Hewitt and Manning(2019)}]{Hewitt:19b}
John Hewitt and Christopher~D. Manning. 2019.
\newblock {A} structural probe for finding syntax in word representations.
\newblock In \emph{Proceedings of the 2019 Conference of the North {A}merican
  Chapter of the Association for Computational Linguistics: Human Language
  Technologies, Volume 1 (Long and Short Papers)}, pages 4129--4138.

\bibitem[{Hochreiter and Schmidhuber(1997)}]{Hochreiter:97}
Sepp Hochreiter and J{\"u}rgen Schmidhuber. 1997.
\newblock Long short-term memory.
\newblock \emph{Neural computation}, 9(8):1735--1780.

\bibitem[{Holtzman et~al.(2018)Holtzman, Buys, Forbes, Bosselut, Golub, and
  Choi}]{Holtzman:18}
Ari Holtzman, Jan Buys, Maxwell Forbes, Antoine Bosselut, David Golub, and
  Yejin Choi. 2018.
\newblock Learning to write with cooperative discriminators.
\newblock In \emph{Proceedings of the 56th Annual Meeting of the Association
  for Computational Linguistics (Volume 1: Long Papers)}, pages 1638--1649.

\bibitem[{Hu et~al.(2020{\natexlab{a}})Hu, Chen, and Levy}]{Hu:20}
Jennifer Hu, Sherry~Y Chen, and Roger~P Levy. 2020{\natexlab{a}}.
\newblock A closer look at the performance of neural language models on
  reflexive anaphor licensing.
\newblock \emph{Proceedings of the Society for Computation in Linguistics},
  3(1):382--392.

\bibitem[{Hu et~al.(2020{\natexlab{b}})Hu, Cheng, Gan, Liu, Gao, and
  Neubig}]{Hu:20b}
Junjie Hu, Yu~Cheng, Zhe Gan, Jingjing Liu, Jianfeng Gao, and Graham Neubig.
  2020{\natexlab{b}}.
\newblock What makes a good story? {D}esigning composite rewards for visual
  storytelling.
\newblock In \emph{Proceedings of the Thirty-Fourth AAAI Conference on
  Artificial Intelligence}, pages 7969--7976.

\bibitem[{Jamshid~Lou et~al.(2018)Jamshid~Lou, Anderson, and Johnson}]{Lou:18}
Paria Jamshid~Lou, Peter Anderson, and Mark Johnson. 2018.
\newblock Disfluency detection using auto-correlational neural networks.
\newblock In \emph{Proceedings of the 2018 Conference on Empirical Methods in
  Natural Language Processing}, pages 4610--4619.

\bibitem[{Ji and Smith(2017)}]{Ji:17}
Yangfeng Ji and Noah~A. Smith. 2017.
\newblock Neural discourse structure for text categorization.
\newblock In \emph{Proceedings of the 55th Annual Meeting of the Association
  for Computational Linguistics (Volume 1: Long Papers)}, pages 996--1005.

\bibitem[{Johnson and Charniak(2004)}]{Johnson:04}
Mark Johnson and Eugene Charniak. 2004.
\newblock A {TAG}-based noisy-channel model of speech repairs.
\newblock In \emph{Proceedings of the 42nd Annual Meeting of the Association
  for Computational Linguistics}, pages 33--39.

\bibitem[{Kiros et~al.(2015)Kiros, Zhu, Salakhutdinov, Zemel, Torralba,
  Urtasun, and Fidler}]{Kiros:2015}
Ryan Kiros, Yukun Zhu, Ruslan Salakhutdinov, Richard~S. Zemel, Antonio
  Torralba, Raquel Urtasun, and Sanja Fidler. 2015.
\newblock Skip-thought vectors.
\newblock In \emph{Proceedings of the 28th International Conference on Neural
  Information Processing Systems - Volume 2}, pages 3294--3302.

\bibitem[{Kuncoro et~al.(2018)Kuncoro, Dyer, Hale, Yogatama, Clark, and
  Blunsom}]{Clark:18}
Adhiguna Kuncoro, Chris Dyer, John Hale, Dani Yogatama, Stephen Clark, and Phil
  Blunsom. 2018.
\newblock {LSTM}s can learn syntax-sensitive dependencies well, but modeling
  structure makes them better.
\newblock In \emph{Proceedings of the 56th Annual Meeting of the Association
  for Computational Linguistics (Volume 1: Long Papers)}, pages 1426--1436.

\bibitem[{Lai and Tetreault(2018)}]{Lai:18}
Alice Lai and Joel Tetreault. 2018.
\newblock Discourse coherence in the wild: {A} dataset, evaluation and methods.
\newblock In \emph{Proceedings of the 19th Annual {SIG}dial Meeting on
  Discourse and Dialogue}, pages 214--223.

\bibitem[{Lan et~al.(2020)Lan, Chen, Goodman, Gimpel, Sharma, and
  Soricut}]{Lan:20}
Zhenzhong Lan, Mingda Chen, Sebastian Goodman, Kevin Gimpel, Piyush Sharma, and
  Radu Soricut. 2020.
\newblock {ALBERT:} {A} lite {BERT} for self-supervised learning of language
  representations.
\newblock In \emph{Proceedings of the 8th International Conference on Learning
  Representations}.

\bibitem[{Lapata(2003)}]{Lapata:03}
Mirella Lapata. 2003.
\newblock Probabilistic text structuring: {E}xperiments with sentence ordering.
\newblock In \emph{Proceedings of the 41st Annual Meeting of the Association
  for Computational Linguistics}, pages 545--552.

\bibitem[{Lau et~al.(2015)Lau, Clark, and Lappin}]{Lau:15}
Jey~Han Lau, Alexander Clark, and Shalom Lappin. 2015.
\newblock Unsupervised prediction of acceptability judgements.
\newblock In \emph{Proceedings of the 53rd Annual Meeting of the Association
  for Computational Linguistics and the 7th International Joint Conference on
  Natural Language Processing (Volume 1: Long Papers)}, pages 1618--1628.

\bibitem[{Li and Jurafsky(2017)}]{Li:17}
Jiwei Li and Dan Jurafsky. 2017.
\newblock Neural net models of open-domain discourse coherence.
\newblock In \emph{Proceedings of the 2017 Conference on Empirical Methods in
  Natural Language Processing}, pages 198--209.

\bibitem[{Li et~al.(2018)Li, Chen, Nie, Liu, Feng, and Cai}]{Li:18}
Xia Li, Minping Chen, Jianyun Nie, Zhenxing Liu, Ziheng Feng, and Yingdan Cai.
  2018.
\newblock Coherence-based automated essay scoring using self-attention.
\newblock In \emph{Proceedings of the 17th China National Conference on
  Computational Linguistics, CCL 2018, and the 6th International Symposium on
  Natural Language Processing Based on Naturally Annotated Big Data}, pages
  386--397.

\bibitem[{Liang et~al.(2018)Liang, Li, Su, Bian, Li, and Shi}]{Liang:18}
Bin Liang, Hongcheng Li, Miaoqiang Su, Pan Bian, Xirong Li, and Wenchang Shi.
  2018.
\newblock Deep text classification can be fooled.
\newblock In \emph{Proceedings of the Twenty-Seventh International Joint
  Conference on Artificial Intelligence}, pages 4208--4215.

\bibitem[{Liu et~al.(2019)Liu, Ott, Goyal, Du, Joshi, Chen, Levy, Lewis,
  Zettlemoyer, and Stoyanov}]{Liu:19}
Yinhan Liu, Myle Ott, Naman Goyal, Jingfei Du, Mandar Joshi, Danqi Chen, Omer
  Levy, Mike Lewis, Luke Zettlemoyer, and Veselin Stoyanov. 2019.
\newblock Roberta: {A} robustly optimized {BERT} pretraining approach.
\newblock \emph{CoRR}, cs.CL/1907.11692v1.

\bibitem[{Mann and Thompson(1988)}]{Mann:88}
William~C Mann and Sandra~A Thompson. 1988.
\newblock Rhetorical structure theory: {T}oward a functional theory of text
  organization.
\newblock \emph{Text}, 8(3):243--281.

\bibitem[{Marvin and Linzen(2018)}]{Marvin:18}
Rebecca Marvin and Tal Linzen. 2018.
\newblock Targeted syntactic evaluation of language models.
\newblock In \emph{Proceedings of the 2018 Conference on Empirical Methods in
  Natural Language Processing}, pages 1192--1202.

\bibitem[{Mave et~al.(2018)Mave, Maharjan, and Solorio}]{Mave:18}
Deepthi Mave, Suraj Maharjan, and Thamar Solorio. 2018.
\newblock Language identification and analysis of code-switched social media
  text.
\newblock In \emph{Proceedings of the Third Workshop on Computational
  Approaches to Linguistic Code-Switching}, pages 51--61.

\bibitem[{McCoy et~al.(2019)McCoy, Pavlick, and Linzen}]{McCoy:19}
Tom McCoy, Ellie Pavlick, and Tal Linzen. 2019.
\newblock Right for the wrong reasons: {D}iagnosing syntactic heuristics in
  natural language inference.
\newblock In \emph{Proceedings of the 57th Annual Meeting of the Association
  for Computational Linguistics}, pages 3428--3448.

\bibitem[{Mesgar and Strube(2016)}]{Mesgar:16}
Mohsen Mesgar and Michael Strube. 2016.
\newblock Lexical coherence graph modeling using word embeddings.
\newblock In \emph{Proceedings of the 2016 Conference of the North {A}merican
  Chapter of the Association for Computational Linguistics: Human Language
  Technologies}, pages 1414--1423.

\bibitem[{Mesgar and Strube(2018)}]{Mesgar:18}
Mohsen Mesgar and Michael Strube. 2018.
\newblock A neural local coherence model for text quality assessment.
\newblock In \emph{Proceedings of the 2018 Conference on Empirical Methods in
  Natural Language Processing}, pages 4328--4339.

\bibitem[{Miltsakaki et~al.(2004)Miltsakaki, Prasad, Joshi, and
  Webber}]{Miltsakaki:04}
Eleni Miltsakaki, Rashmi Prasad, Aravind Joshi, and Bonnie Webber. 2004.
\newblock The {P}enn discourse treebank.
\newblock In \emph{Proceedings of the Fourth International Conference on
  Language Resources and Evaluation}.

\bibitem[{Mim et~al.(2019)Mim, Inoue, Reisert, Ouchi, and Inui}]{Mim:19}
Farjana~Sultana Mim, Naoya Inoue, Paul Reisert, Hiroki Ouchi, and Kentaro Inui.
  2019.
\newblock Unsupervised learning of discourse-aware text representation.
\newblock In \emph{Proceedings of the 2019 Student Research Workshop}, pages
  93--104.

\bibitem[{Nallapati et~al.(2016)Nallapati, Zhou, dos Santos,
  G{\"{u}}l{\c{c}}ehre, and Xiang}]{Nallapati:16}
Ramesh Nallapati, Bowen Zhou, C{\'{\i}}cero~Nogueira dos Santos, {\c{C}}aglar
  G{\"{u}}l{\c{c}}ehre, and Bing Xiang. 2016.
\newblock Abstractive text summarization using sequence-to-sequence {RNN}s and
  beyond.
\newblock In \emph{Proceedings of The 20th {SIGNLL} Conference on Computational
  Natural Language Learning}, pages 280--290.

\bibitem[{Papernot et~al.(2018)Papernot, Faghri, Carlini, Goodfellow, Feinman,
  Kurakin, Xie, Sharma, Brown, Roy, Matyasko, Behzadan, Hambardzumyan, Zhang,
  Juang, Li, Sheatsley, Garg, Uesato, Gierke, Dong, Berthelot, Hendricks,
  Rauber, and Long}]{Papernot:18}
Nicolas Papernot, Fartash Faghri, Nicholas Carlini, Ian Goodfellow, Reuben
  Feinman, Alexey Kurakin, Cihang Xie, Yash Sharma, Tom Brown, Aurko Roy,
  Alexander Matyasko, Vahid Behzadan, Karen Hambardzumyan, Zhishuai Zhang,
  Yi-Lin Juang, Zhi Li, Ryan Sheatsley, Abhibhav Garg, Jonathan Uesato, Willi
  Gierke, Yinpeng Dong, David Berthelot, Paul Hendricks, Jonas Rauber, and
  Rujun Long. 2018.
\newblock Technical report on the {C}lever{H}ans v2.1.0 adversarial examples
  library.
\newblock \emph{CoRR}, cs.LG/1610.00768v6.

\bibitem[{Park and Kim(2015)}]{Park:15}
Cesc~C Park and Gunhee Kim. 2015.
\newblock Expressing an image stream with a sequence of natural sentences.
\newblock In \emph{Proceedings of Advances in Neural Information Processing
  Systems 28}, pages 73--81.

\bibitem[{Parker et~al.(2011)Parker, Graff, Kong, Chen, and Maeda}]{Parker:11}
Robert Parker, David Graff, Junbo Kong, Ke~Chen, and Kazuaki Maeda. 2011.
\newblock English gigaword fifth edition, linguistic data consortium.
\newblock \emph{Google Scholar}.

\bibitem[{Pennington et~al.(2014)Pennington, Socher, and
  Manning}]{Pennington:14}
Jeffrey Pennington, Richard Socher, and Christopher Manning. 2014.
\newblock {G}lo{V}e: {G}lobal vectors for word representation.
\newblock In \emph{Proceedings of the 2014 Conference on Empirical Methods in
  Natural Language Processing}, pages 1532--1543.

\bibitem[{Peters et~al.(2018)Peters, Neumann, Zettlemoyer, and Yih}]{Peters:18}
Matthew Peters, Mark Neumann, Luke Zettlemoyer, and Wen-tau Yih. 2018.
\newblock Dissecting contextual word embeddings: {A}rchitecture and
  representation.
\newblock In \emph{Proceedings of the 2018 Conference on Empirical Methods in
  Natural Language Processing}, pages 1499--1509.

\bibitem[{Pitler et~al.(2010)Pitler, Louis, and Nenkova}]{Pitler:10}
Emily Pitler, Annie Louis, and Ani Nenkova. 2010.
\newblock Automatic evaluation of linguistic quality in multi-document
  summarization.
\newblock In \emph{Proceedings of the 48th Annual Meeting of the Association
  for Computational Linguistics}, pages 544--554.

\bibitem[{Pitler and Nenkova(2008)}]{Pitler:08}
Emily Pitler and Ani Nenkova. 2008.
\newblock Revisiting readability: {A} unified framework for predicting text
  quality.
\newblock In \emph{Proceedings of the 2008 Conference on Empirical Methods in
  Natural Language Processing}, pages 186--195.

\bibitem[{Prabhumoye et~al.(2020)Prabhumoye, Salakhutdinov, and
  Black}]{Prabhumoye:20}
Shrimai Prabhumoye, Ruslan Salakhutdinov, and Alan~W Black. 2020.
\newblock Topological sort for sentence ordering.
\newblock In \emph{Proceedings of the 58th Annual Meeting of the Association
  for Computational Linguistics}, pages 2783--2792.

\bibitem[{Prasad et~al.(2008)Prasad, Dinesh, Lee, Miltsakaki, Robaldo, Joshi,
  and Webber}]{Prasad:08}
Rashmi Prasad, Nikhil Dinesh, Alan Lee, Eleni Miltsakaki, Livio Robaldo,
  Aravind Joshi, and Bonnie Webber. 2008.
\newblock The {P}enn discourse {T}ree{B}ank 2.0.
\newblock In \emph{Proceedings of the Sixth International Conference on
  Language Resources and Evaluation}.

\bibitem[{Putra and Tokunaga(2017)}]{Putra:17}
Jan Wira~Gotama Putra and Takenobu Tokunaga. 2017.
\newblock Evaluating text coherence based on semantic similarity graph.
\newblock In \emph{Proceedings of {T}ext{G}raphs-11: the Workshop on
  Graph-based Methods for Natural Language Processing}, pages 76--85.

\bibitem[{Samanta and Mehta(2017)}]{Samanta:17}
Suranjana Samanta and Sameep Mehta. 2017.
\newblock Towards crafting text adversarial samples.
\newblock \emph{CoRR}, cs.LG/1707.02812v1.

\bibitem[{Sato et~al.(2018)Sato, Suzuki, Shindo, and Matsumoto}]{Sato:18}
Motoki Sato, Jun Suzuki, Hiroyuki Shindo, and Yuji Matsumoto. 2018.
\newblock Interpretable adversarial perturbation in input embedding space for
  text.
\newblock In \emph{Proceedings of the Twenty-Seventh International Joint
  Conference on Artificial Intelligence}, pages 4323--4330.

\bibitem[{Somasundaran et~al.(2014)Somasundaran, Burstein, and
  Chodorow}]{Somasundaran:14}
Swapna Somasundaran, Jill Burstein, and Martin Chodorow. 2014.
\newblock Lexical chaining for measuring discourse coherence quality in
  test-taker essays.
\newblock In \emph{Proceedings of the 25th International Conference on
  Computational Linguistics: Technical Papers}, pages 950--961.

\bibitem[{Tang et~al.(2018)Tang, M{\"u}ller, Rios, and Sennrich}]{Tang:18}
Gongbo Tang, Mathias M{\"u}ller, Annette Rios, and Rico Sennrich. 2018.
\newblock Why self-attention? {A} targeted evaluation of neural machine
  translation architectures.
\newblock In \emph{Proceedings of the 2018 Conference on Empirical Methods in
  Natural Language Processing}, pages 4263--4272.

\bibitem[{Tay et~al.(2018)Tay, Phan, Tuan, and Hui}]{Tay:18}
Yi~Tay, Minh~C. Phan, Luu~Anh Tuan, and Siu~Cheung Hui. 2018.
\newblock {SkipFlow}: {I}ncorporating neural coherence features for end-to-end
  automatic text scoring.
\newblock In \emph{Proceedings of the Thirty-Second AAAI Conference on
  Artificial Intelligence}, pages 5948--5955.

\bibitem[{Tien~Nguyen and Joty(2017)}]{Nguyen:17}
Dat Tien~Nguyen and Shafiq Joty. 2017.
\newblock A neural local coherence model.
\newblock In \emph{Proceedings of the 55th Annual Meeting of the Association
  for Computational Linguistics (Volume 1: Long Papers)}, pages 1320--1330.

\bibitem[{Tran et~al.(2018)Tran, Bisazza, and Monz}]{Tran:18}
Ke~Tran, Arianna Bisazza, and Christof Monz. 2018.
\newblock The importance of being recurrent for modeling hierarchical
  structure.
\newblock In \emph{Proceedings of the 2018 Conference on Empirical Methods in
  Natural Language Processing}, pages 4731--4736.

\bibitem[{Trinh and Le(2018)}]{Trinh:18}
Trieu~H. Trinh and Quoc~V. Le. 2018.
\newblock A simple method for commonsense reasoning.
\newblock \emph{CoRR}, cs.AI/1806.02847v2.

\bibitem[{Warstadt et~al.(2019)Warstadt, Singh, and Bowman}]{Warstadt:19}
Alex Warstadt, Amanpreet Singh, and Samuel Bowman. 2019.
\newblock Neural network acceptability judgments.
\newblock \emph{Transactions of the Association for Computational Linguistics},
  7(0).

\bibitem[{Webber(2009)}]{Webber:09}
Bonnie Webber. 2009.
\newblock Genre distinctions for discourse in the {P}enn {T}ree{B}ank.
\newblock In \emph{Proceedings of the Joint Conference of the 47th Annual
  Meeting of the ACL and the 4th International Joint Conference on Natural
  Language Processing of the AFNLP}, pages 674--682.

\bibitem[{Wilcox et~al.(2018)Wilcox, Levy, Morita, and Futrell}]{Wilcox:18}
Ethan Wilcox, Roger Levy, Takashi Morita, and Richard Futrell. 2018.
\newblock What do {RNN} language models learn about filler{--}gap dependencies?
\newblock In \emph{Proceedings of the 2018 {EMNLP} Workshop {B}lackbox{NLP}:
  Analyzing and Interpreting Neural Networks for {NLP}}, pages 211--221.

\bibitem[{Wilson and Sperber(2004)}]{Wilson:04}
Deirdre Wilson and Dan Sperber. 2004.
\newblock Relevance theory.
\newblock In \emph{The Handbook of Pragmatics}, pages 607--632. Blackwell.

\bibitem[{Xu et~al.(2019)Xu, Saghir, Kang, Long, Bose, Cao, and Cheung}]{Xu:19}
Peng Xu, Hamidreza Saghir, Jin~Sung Kang, Teng Long, Avishek~Joey Bose,
  Yanshuai Cao, and Jackie Chi~Kit Cheung. 2019.
\newblock A cross-domain transferable neural coherence model.
\newblock In \emph{Proceedings of the 57th Annual Meeting of the Association
  for Computational Linguistics}, pages 678--687.

\bibitem[{Yang et~al.(2020)Yang, Chen, Hsieh, Wang, and Jordan}]{Yang:18b}
Puyudi Yang, Jianbo Chen, Cho{-}Jui Hsieh, Jane{-}Ling Wang, and Michael~I.
  Jordan. 2020.
\newblock Greedy attack and gumbel attack: {G}enerating adversarial examples
  for discrete data.
\newblock \emph{Journal of Machine Learning Research}, 21:43:1--43:36.

\bibitem[{Yang et~al.(2015)Yang, Yih, and Meek}]{Yang:15}
Yi~Yang, Wen-tau Yih, and Christopher Meek. 2015.
\newblock {W}iki{QA}: {A} challenge dataset for open-domain question answering.
\newblock In \emph{Proceedings of the 2015 Conference on Empirical Methods in
  Natural Language Processing}, pages 2013--2018.

\bibitem[{Yang et~al.(2019)Yang, Dai, Yang, Carbonell, Salakhutdinov, and
  Le}]{Yang:19}
Zhilin Yang, Zihang Dai, Yiming Yang, Jaime Carbonell, Russ~R Salakhutdinov,
  and Quoc~V Le. 2019.
\newblock Xlnet: {G}eneralized autoregressive pretraining for language
  understanding.
\newblock In \emph{Proceedings of the Thirty-third Conference on Neural
  Information Processing Systems}, pages 5754--5764.

\bibitem[{Yang et~al.(2018)Yang, Qi, Zhang, Bengio, Cohen, Salakhutdinov, and
  Manning}]{Yang:18}
Zhilin Yang, Peng Qi, Saizheng Zhang, Yoshua Bengio, William Cohen, Ruslan
  Salakhutdinov, and Christopher~D. Manning. 2018.
\newblock {H}otpot{QA}: {A} dataset for diverse, explainable multi-hop question
  answering.
\newblock In \emph{Proceedings of the 2018 Conference on Empirical Methods in
  Natural Language Processing}, pages 2369--2380.

\bibitem[{Yirmibe{\c{s}}o{\u{g}}lu and Eryi{\u{g}}it(2018)}]{Yirmibesoglu:18}
Zeynep Yirmibe{\c{s}}o{\u{g}}lu and G{\"u}l{\c{s}}en Eryi{\u{g}}it. 2018.
\newblock Detecting code-switching between {T}urkish-{E}nglish language pair.
\newblock In \emph{Proceedings of the 2018 {EMNLP} Workshop W-{NUT}: The 4th
  Workshop on Noisy User-generated Text}, pages 110--115.

\bibitem[{Zellers et~al.(2018)Zellers, Bisk, Schwartz, and Choi}]{Zellers:18}
Rowan Zellers, Yonatan Bisk, Roy Schwartz, and Yejin Choi. 2018.
\newblock {SWAG}: {A} large-scale adversarial dataset for grounded commonsense
  inference.
\newblock In \emph{Proceedings of the 2018 Conference on Empirical Methods in
  Natural Language Processing}, pages 93--104.

\bibitem[{Zellers et~al.(2019)Zellers, Holtzman, Bisk, Farhadi, and
  Choi}]{Zellers:19}
Rowan Zellers, Ari Holtzman, Yonatan Bisk, Ali Farhadi, and Yejin Choi. 2019.
\newblock {H}ella{S}wag: {C}an a machine really finish your sentence?
\newblock In \emph{Proceedings of the 57th Annual Meeting of the Association
  for Computational Linguistics}, pages 4791--4800.

\bibitem[{Zhu et~al.(2015)Zhu, Kiros, Zemel, Salakhutdinov, Urtasun, Torralba,
  and Fidler}]{Zhu:15}
Yukun Zhu, Ryan Kiros, Richard~S. Zemel, Ruslan Salakhutdinov, Raquel Urtasun,
  Antonio Torralba, and Sanja Fidler. 2015.
\newblock Aligning books and movies: {T}owards story-like visual explanations
  by watching movies and reading books.
\newblock In \emph{Proceedings of the 2015 IEEE Conference on Computer Vision
  and Pattern Recognition}, pages 19--27.

\end{thebibliography}

\end{document}
